\documentclass[11pt]{article}

\usepackage{acl}

\usepackage{times}
\usepackage{latexsym}
\usepackage{booktabs}
\usepackage{multirow,multicol}
\usepackage{subfigure}
\usepackage[table]{xcolor}
\usepackage{colortbl}
\usepackage{amsmath,amsfonts}

\usepackage[T1]{fontenc}

\usepackage[utf8]{inputenc}

\usepackage{microtype}

\usepackage{inconsolata}

\usepackage{graphicx}

\usepackage{algorithm}
\usepackage{algpseudocode}
\usepackage{hyperref}

\newtheorem{definition}{Definition}

\title{ThinkProbe: Beyond Accuracy - Structural Profiling of Open-Ended LLM Reasoning Traces via Non-Generative Thought Graphs\thanks{The code will be availble at : \href{https://github.com/kmamine/ThinkProb}{https://github.com/kmamine/ThinkProb} }}

\author{
  \textbf{Mohamed Amine Kerkouri\textsuperscript{1}},
  \textbf{Simon D. Hernandez\textsuperscript{1}},
  \textbf{Marouane Tliba\textsuperscript{2}},
\\
  \textbf{Yann Dauxais\textsuperscript{1}},
  \textbf{Maha Ben-Fares\textsuperscript{1}},
  \textbf{Pierre Holat\textsuperscript{1}}
\\
\\
  \textsuperscript{1}F-Initiatives, Paris, France \quad
  \textsuperscript{2}Université sorbonne Paris Nord, Villetaneuse
, France
\\
}

\begin{document}
\maketitle
\begin{abstract}
We present \textbf{ThinkProbe}, a framework for structural analysis of
LLM reasoning traces.
ThinkProbe converts each trace into a \textbf{Thought Graph} a
directed graph with cycles, 8 node types, and 6 edge types and
derives a \textbf{19-metric five-dimensional cognitive profile}
(5D-CP: Breadth, Depth, Structure, Metacognitive, Efficiency) through
a fully non-generative pipeline combining rule-based segmentation and
discriminative semantic linking.
Applied to 4{,}200 traces from 7 native reasoning models across 200
open-ended questions and 10 cognitive domains, ThinkProbe reveals
that reasoning structure is a stable, model-level property:
between-model variance exceeds between-domain variance by up to fourfold across four of five cognitive dimensions, with Structure showing genuine sensitivity to question domain, exposing qualitatively distinct cognitive
profiles invisible to accuracy-based evaluation.
\end{abstract}

\section{Introduction}

Existing benchmarks evaluate large language models (LLMs) on \emph{what} they
answer: a scalar accuracy on closed tasks such as MATH~\cite{hendrycksmath2021},
MMLU~\cite{hendrycks2021mmlu}, or GSM8K~\cite{cobbe2021gsm8k}.
This paradigm is adequate when outputs were short and deterministic.
It is no longer sufficient, in the more generic different uses contexts of LLMs.
Contemporary reasoning models generate thousands of tokens of internal
deliberation inside \texttt{<think>} tags before producing a final
answer.
This pre-answer thinking is trained via outcome-based reinforcement,
rather than token-level supervision, its structure reflects emergent
cognitive behaviour, not learned output formatting, yet no framework
exists to characterise it systematically across models and domains.

This accuracy-centric paradigm rests on a foundational assumption that is increasingly untenable: the existence of a ground truth. For a growing class of real-world tasks — ethical dilemmas, philosophical inquiry, open-ended ideation, strategic planning under uncertainty — no correct answer exists. A model asked to propose novel startup ideas, reason through a moral conflict, or evaluate competing geopolitical scenarios, cannot be assessed by comparing its output to a reference. Yet these are precisely the tasks for which frontier reasoning models are deployed. In the absence of a correctness signal, the reasoning process itself becomes the only observable proxy for quality.

We introduce \textbf{ThinkProbe}, a framework that fills this gap. ThinkProbe extracts a \textbf{Thought Graph}, a directed graph with cycles, 8 node types, and 6 edge types via a fully non-generative pipeline, and derives a 19-metric \textbf{five-dimensional cognitive profile} (5D-CP) spanning Breadth, Depth, Structure, Metacognitive, and Efficiency.

Beyond characterisation, ThinkProbe opens three practical directions: real-time structural confidence signals during inference, outlier detection for reasoning failures invisible at output level, and profile-aware model selection matching cognitive style to task demands.

Our contributions are the following:
\begin{itemize}
\itemsep0em
  \item[\textbf{C1}] A \textbf{cycle-permitting Thought Graph}
    representation capturing backtracking, synthesis, and cross-branch
    connectivity phenomena invisible to Directed Acyclic graphs (DAGs) and tree approaches.
  \item[\textbf{C2}] A \textbf{fully non-generative extraction
    pipeline} combining rule-based segmentation, MiniLM-based TextTiling, and cross-segment semantic linking, eliminating LLM-analysing-LLM circularity.
  \item[\textbf{C3}] A \textbf{19-metric 5D cognitive profile}
    statistically validated across 4{,}200 traces: all metrics
    discriminate models ($\varepsilon^2 = 0.10$--$0.75$,
    all $p < 0.001$).
  \item[\textbf{C4}] An \textbf{empirical study across 7 native reasoning models, 200 open-ended questions, and 10 cognitive domains}, showing that reasoning structure is a stable model-level property across four of five cognitive dimensions, with between-model variance exceeding between-domain variance by up  fourfold on most dimensions.
\end{itemize}

The remainder of the paper describes the framework (§\ref{sec:method}),
experimental protocol (§\ref{sec:protocol}), results
(§\ref{sec:results}), conclusion (§\ref{section:conclusion}), and limitations (§\ref{sec:limitations}).

\section{Related Work}
\label{sec:related}

\paragraph{Graph-based trace analysis.}
Chain-of-Thought prompting~\cite{wei2022chain} and its
extensions Tree of Thoughts~\cite{yao2023tree} and self-consistency~\cite{wang2023selfconsistency} established
structured generation as a path to improved reasoning, but treat
structure as an \emph{input} to generation rather than a property
to be measured in intrinsically occurring traces. Graph of Thoughts~\cite{besta2024got} models LLM outputs as arbitrary
graphs at inference time to improve generation quality, but does not
analyse naturally occurring traces.
Mapping the Minds of LLMs~\cite{xiong2025mapping} builds directed
reasoning graphs from CoT outputs and shows that branching and
convergence ratios correlate with accuracy.
ReasoningFlow~\cite{lee2025reasoningflow} parses traces into DAGs to
characterise reasoning patterns as subgraph structures.
LCoT2Tree~\cite{jiang2025LCOTTree} converts long chains into
hierarchical trees, finding that structural patterns predict
task performance.
CoTJudger~\cite{li2026cotjudger} extracts dependency graphs and
identifies the Shortest Effective Path to quantify redundancy.
Three limitations cut across these works: DAG and tree representations
cannot encode cycles (iterative refinement and perspective oscillation
are real trace phenomena); extraction relies on LLM-based clustering
or conversion; and the evaluation target is accuracy or efficiency on
closed tasks, not open-ended cognitive profiling.

\paragraph{Cognitive profiling.}
CogBench~\cite{coda2024cogbench} derives ten behavioural metrics from seven cognitive psychology experiments applied to closed tasks.
Cognitive Foundations for reasoning~\cite{kargupta2025cognitive}
proposes a 28-element taxonomy and analyses 170K traces from 17 models, finding that models under-utilise meta-cognitive elements on ill-structured problems.
CogTest~\cite{dong2025cogtest} evaluates 16 Habits of Mind, showing that reasoning models deploy human-like habits adaptively across tasks.
ThinkARM~\cite{li2025schoenfeld} applies Schoenfeld's Episode Theory to abstract traces into functional steps(Analysis, Explore, Implement, Verify, and Watch)revealing reproducible thinking dynamics across models.
MetaCog-Bench~\cite{anonymous2026metacogbench}\footnote{paper under review in openreview (authors hidden by the system).} benchmarks metacognitive monitoring and control in LLMs.
Earlier step-level evaluation work, including ROSCOE~\cite{golovneva2023roscoe} for coherence and faithfulness scoring and process reward models~\cite{lightman2024verify} for step correctness, targets quality of individual reasoning steps rather than holistic structural profiles.
Taken together, these works either depend on closed-task correctness signals, require LLM-assisted annotation, or address single cognitive dimensions in isolation.
We address all three limitations simultaneously, unifying Breadth, Depth, Structure, Metacognitive, and Efficiency into a single non-generative profile applicable to open-ended reasoning.

\paragraph{Efficiency and overthinking.}
THINK-Bench~\cite{li2025think} evaluates thinking efficiency and
chain-of-thought quality in large reasoning models, finding that most
models overthink on simpler tasks.
OptimalThinkingBench~\cite{aggarwal2025optimalthinkingbench} jointly
benchmarks over- and under-thinking as accuracy--token trade-offs.
TRACE~\cite{zhang2025trace} decomposes traces into Explorer and
Late-Landing patterns to diagnose verbosity.
ThinkProbe treats efficiency as one dimension of a
five-dimensional profile rather than a primary evaluation target.

\section{Method}
\label{sec:method}

\begin{figure*}
    \centering
    \includegraphics[width=0.7\textwidth]{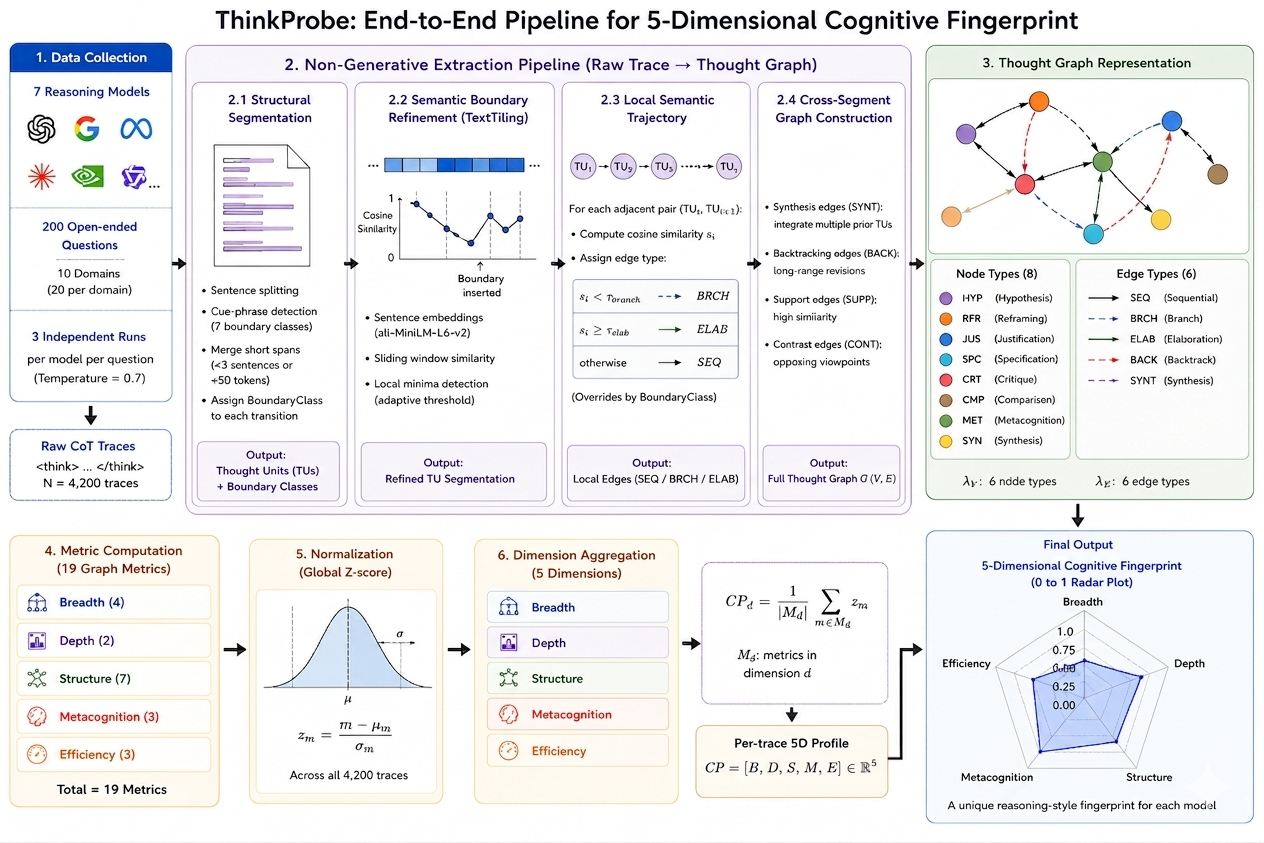}
    \caption{ThinkProbe pipeline}
    \label{fig:general-pipeline}
    \vspace{-5mm}
\end{figure*}

\textbf{ThinkProbe} takes a chain-of-thought trace as input and produces a \textbf{5D Cognitive Profile (5D-CP)}, a vector in $\mathbb{R}^5$ characterizing a model's reasoning style along five interpretable dimensions: Breadth, Depth, Structure, Metacognition, and Efficiency.
Figure~\ref{fig:general-pipeline} illustrates the full system.
The trace is first segmented into \textit{Thought Units} (TUs), contiguous text spans each expressing a single cognitive move by a layered pipeline that uses no generative model.
TUs are then linked into a directed \textit{Thought Graph}, and 19 behavioral metrics are computed from the graph structure.
Finally, the metrics are aggregated into the 5D-CP via global z-score normalization.
The generative-model-free design is deliberate: using a secondary LLM to analyze a primary LLM's output imports the analyzer's own biases and reasoning tendencies, making the measurement circular.
All extraction is either rule-based (structural segmentation, node classification) or embedding-based (boundary refinement and edge typing).

\subsection{5D Cognitive Profile}
\label{sec:method:5dcp}

The 5D Cognitive Profile (5D-CP) is grounded in a simple observation about how models
approach open-ended problems: some explore a wide range of ideas
while others pursue a narrower set in greater depth.
This breadth--depth distinction is the conceptual seed of the
framework, extended to a five-dimensional profile that together
captures the full picture of reasoning style:

\begin{itemize}
\itemsep0em
  \item \textbf{Breadth}: how widely does the model explore the
        idea space? Does it generate diverse hypotheses, perspectives,
        and angles before committing?
  \item \textbf{Depth}: how deeply does it elaborate within a line
        of reasoning? Does it build extended chains of justification
        and specification?
  \item \textbf{Structure}: what connective behaviors link breadth
        and depth? Does the model backtrack, synthesize across
        branches, and converge toward conclusions?
  \item \textbf{Metacognition}: how self-aware is the reasoning?
        Does the model critique its own ideas, hedge uncertainty,
        and adopt alternative perspectives?
  \item \textbf{Efficiency}: how economically does the model use
        its token budget? Does it elaborate concisely or verbosely
        relative to the number of ideas it produces?
\end{itemize}

These dimensions are not mutually exclusive; they are complementary
lenses on the same reasoning process.
A model with high Breadth and low Depth is a wide-but-shallow explorer;
one with high Depth and high Efficiency is a focused, economical deliberator.
The \textbf{5D-CP} makes these patterns explicit, comparable, and independent
of whether any answer is correct, a property that makes it particularly
suited to open-ended domains where no ground truth exists.

\subsection{Thought Graph}
\label{sec:method:tg}

A \textbf{Thought Graph} is the formal representation from which
all \textbf{ThinkProbe} metrics are computed.

\begin{definition}[Thought Graph]
A Thought Graph $G = (V, E, \lambda_V, \lambda_E)$ is a directed
graph where:
\begin{itemize}
\itemsep0em
  \item $V = \{v_0, \ldots, v_n\}$ is an ordered set of
        \textbf{Thought Units} (TUs) contiguous text segments
        each expressing one cognitive move;
  \item $E \subseteq V \times V$ is a set of directed edges,
        permitting cycles;
  \item $\lambda_V : V \to \mathcal{N}$ assigns each TU a node
        type from an 8-type taxonomy (\S\ref{sec:method:taxonomy});
  \item $\lambda_E : E \to \mathcal{E}$ assigns each edge a
        semantic relation from an 6-type taxonomy
        (\S\ref{sec:method:taxonomy}).
\end{itemize}
\end{definition}

Figure~\ref{fig:example_graph} in Appendix~\ref{app:figures}
shows a concrete example Thought Graph extracted from a Gemma-4-31B trace 
on the question ``Should police use predictive policing AI?'' (Ethical 
Dilemmas domain), illustrating the node types, edge types, and long-range 
\textsc{back} arcs that motivate the directed-graph representation.

\paragraph{Why directed graphs with cycles?}
Tree-based representations~\cite{jiang2025LCOTTree} cannot express
cross-branch synthesis; a node that draws from two independent
reasoning branches.
DAG-based representations forbid cycles, but two phenomena in
extended reasoning require them.
First, \emph{convergence}: a synthesis node~$v_j$ that draws from
multiple prior segments~$i_1, i_2 < j$ creates a directed cycle of
length $(j - i + 1)$ through the sequential backbone, a structure
that neither a tree nor a DAG can represent without node duplication.
Second, \emph{iterative refinement}: a model may propose an idea,
critique it, and revise it; this is captured by \textsc{back} edges
and measured independently via backtracking rate and revision depth.
The full directed graph preserves both patterns.
For depth metrics, which require acyclicity, we operate on the
elaboration-edge subgraph $G_{\text{elab}} = (V, E_{\text{ELAB}})$,
which is acyclic by construction.

\paragraph{Structural invariants.}
Every valid Thought Graph satisfies:
(i)~no self-loops;
(ii)~all edge endpoints reference valid TU indices;
(iii)~$G_{\text{elab}}$ is acyclic;
(iv)~every node is reachable from $v_0$;
(v)~at least one Exploration-family node exists.
Traces that cannot satisfy these invariants after extraction
are excluded from analysis ($< 0.2\%$ of all traces).

\subsection{Thought Units and Edges}
\label{sec:method:taxonomy}

\paragraph{Node types.}
Each TU is assigned one of 8 node types organized into four
cognitive families, grounded in the exploration--exploitation
distinction~\citep{march1991exploration}:

\begin{table}[h]
\centering
\small
\setlength{\tabcolsep}{5pt}
\renewcommand{\arraystretch}{1.0}
\begin{tabular}{@{}l l p{3.5cm}@{}}
\toprule
\textbf{Family} & \textbf{Types} & \textbf{Cognitive function} \\
\midrule
Exploration & HYP, RFR      & Generate hypotheses, reframe
                               the problem from a new angle \\
Elaboration & JUS, SPC      & Justify claims, specify details
                               and sub-components \\
Evaluation  & CRT, CMP, MET & Critique, compare alternatives,
                               meta-reflect \\
Convergence & SYN           & Synthesize, and conclude \\
\bottomrule
\end{tabular}
\caption{Node type taxonomy: 8 types in four cognitive families.}
\label{tab:taxonomy}
\vspace{-5mm}
\end{table}

\paragraph{Edge types.}
Six edge types encode the semantic relation between a source 
and target TU: SEQ (default sequential flow), BRCH (branch to a new direction), ELAB (elaboration of the source), BACK (backtrack or revision), SYNT (synthesis across branches), and CRIT (direct critique).
Sequential edges (\textsc{seq}) represent the default flow between
adjacent TUs with no specific semantic relation.
All remaining edges carry an explicit semantic relation and together
form the \textit{semantic edge set} $E_{\text{sem}} = E \setminus E_{\textsc{seq}}$,
used in structural metrics.

\subsection{Extraction Pipeline}
\label{sec:method:pipeline}

The extraction pipeline converts a raw CoT trace into a Thought Graph
through four sequential layers, progressing from surface formatting
cues to semantic connectivity.

\paragraph{Layer 1: Structural segmentation and boundary classification.}
The first layer establishes hard TU boundaries from formatting alone.
Markdown headers (\texttt{\#} to \texttt{\#\#\#\#\#\#}) and bold
standalone lines serve as explicit section breaks and receive the
\textsc{branch} boundary class.
List items produce one TU per item with class \textsc{none}.
Blank lines and single-newline paragraph breaks where the preceding
line ends a sentence and the following begins with an uppercase
letter serve as paragraph separators.

Cue-phrase detection is then applied to the first sentence of each
\textsc{none}-class span to assign supplementary boundary classes:
\textsc{meta} (12 patterns, e.g.\ \textit{let me step back},
\textit{I'm going in circles}),
\textsc{convergence} (16 patterns, e.g.\ \textit{putting this together},
\textit{in summary}), and
\textsc{backtrack} (12 patterns, e.g.\ \textit{wait},
\textit{but actually}, \textit{on second thought}).

Finally, short spans are merged: a \textsc{branch} header of fewer
than 12 tokens absorbs its following span; a \textsc{none} span of
fewer than two sentences or 30 tokens is merged forward into the
adjacent \textsc{none} span.

\paragraph{Layer 2: Soft boundary detection.}
Long \textsc{none}-class spans are further subdivided using a
\textit{TextTiling} procedure with sentence
embeddings from \texttt{all-MiniLM-L6-v2}~\citep{reimers2019sentence, wang2020minilm, allminilml6v2}.
A sliding window of width 3 computes cosine similarity between the
left and right half-windows at each interior sentence position.
Local minima with prominence $\geq 0.15$ and falling below an adaptive
threshold $\tau$ are promoted to new TU boundaries.
$\tau$ is set to the 30th percentile of within-trace inter-sentence
cosine similarities, calibrating the detector to each trace's own
semantic register rather than a fixed global cutoff. Per-trace percentile thresholds self-calibrate to each trace's own semantic register, providing robustness to absolute similarity scale across embedding models and trace lengths.

\paragraph{Layer 3: Semantic trajectory.}
All TU texts are encoded with \texttt{all-MiniLM-L6-v2}.
For each consecutive pair $(v_{i-1}, v_i)$, the cosine similarity
$s_i$ is computed.
Two per-trace thresholds are derived:
$\tau_{\text{branch}} = \text{Pct}_{25}(\{s_i\})$ and
$\tau_{\text{elab}} = \text{Pct}_{65}(\{s_i\})$.
A transition with $s_i < \tau_{\text{branch}}$ receives a \textsc{brch}
edge; one with $s_i \geq \tau_{\text{elab}}$ receives an \textsc{elab}
edge; all others receive \textsc{seq}. Because $\tau_{\text{branch}}$ and $\tau_{\text{elab}}$ are 
derived from the per-trace similarity distribution, boundary 
decisions are invariant to monotone rescaling of the embedding 
space, and the primary discriminability findings ( based on 
rank-order statistics) are robust to embedding model choice 
by construction.

Structural and cue-phrase overrides take priority over cosine
classification: \textsc{branch}-class TUs always emit \textsc{brch};
\textsc{convergence}-class TUs emit \textsc{synt};
\textsc{backtrack}-class TUs emit \textsc{back}.
Each \textsc{brch} boundary increments a segment counter that
partitions the trace into thematic segments for the next layer.

\paragraph{Layer 4: Cross-segment analysis.}
This layer adds non-local edges that span thematic segments.

\textit{Synthesis arcs.}
For each TU $v_j$, if its embedding exceeds cosine similarity $0.50$
to the centroid of at least two prior segments, \textsc{synt} edges
are emitted from $v_j$ to the representative TU of each bridged
segment, and $v_j$'s boundary class is promoted to \textsc{convergence}.

\textit{Revision arcs.}
For each TU $v_j$ with $j > 4$, we search for the highest-similarity
prior TU $v_i$ satisfying:
(i)~$v_i$ belongs to a different segment;
(ii)~$j - i \geq 4$;
(iii)~at least one TU in the interval $(i,\, i{+}4)$ has cosine
similarity below $0.45$ to $v_i$, confirming a topic divergence
between the two.
If such a $v_i$ exists with similarity above $0.65$, a \textsc{back}
edge is emitted from $v_j$ to $v_i$.

\paragraph{Node classification.}
After the four extraction layers, each TU is assigned a node type
by a deterministic priority hierarchy:
(1)~outgoing \textsc{synt} edges to $\geq\!2$ targets $\to$ \textbf{SYN};
(2)~any outgoing \textsc{back} edge $\to$ \textbf{CRT};
(3)~boundary class \textsc{meta} $\to$ \textbf{MET};
(4)~incoming \textsc{brch} and $0.25 < \cos(\mathbf{e}_v, \mathbf{e}_{v_0}) < 0.72$ $\to$ \textbf{RFR};
(5)~incoming \textsc{brch} otherwise $\to$ \textbf{HYP};
(6)~boundary class \textsc{contrast} $\to$ \textbf{CMP};
(7)~boundary class \textsc{support} $\to$ \textbf{JUS};
(8)~boundary class \textsc{convergence} $\to$ \textbf{SYN};
(9)~default: $v_0 \to$ \textbf{HYP}; all others $\to$ \textbf{SPC}.
Rules~4--5 implement the RFR--HYP distinction: a branch-starting TU
that remains semantically anchored to the original problem is a
reframing; one that diverges more sharply opens a new hypothesis.

\subsection{Cognitive Metrics and Profile Construction}
\label{sec:method:metrics}

ThinkProbe computes 19 behavioral metrics from each Thought Graph, organized into the five dimensions of the 5D-CP (Table~\ref{tab:metrics}).
Formal definitions of all metrics appear in Appendix~\ref{app:metrics}.

\begin{table}[t]
\centering
\small
\setlength{\tabcolsep}{4pt}
\renewcommand{\arraystretch}{1.1}
\begin{tabular}{@{}l l p{4.6cm}@{}}
\toprule
\textbf{Dim.} & \textbf{Abbr.} & \textbf{Description} \\
\midrule
\multirow{4}{*}{Breadth}
  & BF  & Proportion of \textsc{brch} edges per node \\
  & UPC & Count of RFR nodes \\
  & DS  & Semantic cluster count among HYP nodes \\
  & FID & Mean pairwise cosine distance among first
          three HYP nodes \\[2pt]
\multirow{2}{*}{Depth}
  & MEC & Longest directed path in $G_{\text{elab}}$ \\
  & MBD & Mean shortest-path depth from source nodes
          in $E_{\text{sem}}$ subgraph \\[2pt]
\multirow{6}{*}{Structure}
  & EER & Exploration-to-Elaboration node ratio \\
  & BR  & Fraction of semantic edges that are \textsc{back} \\
  & CBC & Fraction of branch pairs joined by $\geq\!1$
          \textsc{synt} edge \\
  & CI  & SYN in-degree concentration vs.\ graph mean \\
  & GD  & $|E_{\text{sem}}|/(|V|(|V|-1))$ \\
  & RvD & Mean positional gap over \textsc{back} endpoints \\[2pt]
\multirow{3}{*}{Metacog.}
  & CHR & CRT-to-HYP node ratio \\
  & HD  & Fraction of TUs with uncertainty markers \\
  & PT  & Fraction of nodes that are RFR \\[2pt]
\multirow{4}{*}{Efficiency}
  & TPI & Total tokens\,/\,$\max(\text{UPC},1)$ \\
  & RR  & Fraction of TU pairs with cosine sim.\,$> 0.75$ \\
  & AT  & Mean token count per trace \\
  & ATU & Mean TU count per trace \\
\bottomrule
\end{tabular}
\caption{The 19 active ThinkProbe metrics grouped by 5D dimension.
         Formal definitions in Appendix~\ref{app:metrics}.}
\label{tab:metrics}
\vspace{-5mm}
\end{table}

\paragraph{Breadth.}
\textit{Branching Factor} (BF) measures how frequently the trace
opens new directions, as the proportion of \textsc{brch} edges per node.
\textit{Unique Perspectives} (UPC) counts RFR nodes, the number of
times the model reframes the original problem from a new angle.
\textit{Domain Spread} (DS) quantifies the semantic breadth of
hypotheses by agglomeratively clustering HYP node embeddings
(cosine threshold 0.45) and counting the resulting clusters.
\textit{First-Idea Diversity} (FID) captures how distinct the model's
opening ideas are, measured as mean pairwise cosine distance among
the first three HYP nodes.

\paragraph{Depth.}
\textit{Max Elaboration Chain} (MEC) is the length of the longest
directed path in $G_{\text{elab}}$, measuring how far the model
sustains a single line of reasoning.
\textit{Mean Branch Depth} (MBD) is the mean shortest-path distance
from source nodes to all other nodes in the semantic subgraph,
capturing average elaboration depth across all branches.

\paragraph{Structure.}
\textit{Exploration/Exploitation Ratio} (EER) captures the balance
between generative and elaborative reasoning.
\textit{Backtracking Rate} (BR) measures the fraction of semantic
edges that are revision arcs.
\textit{Cross-Branch Connectivity} (CBC) quantifies integration as
the fraction of branch pairs joined by at least one synthesis.
\textit{Convergence Index} (CI) measures how strongly synthesis
activity concentrates at SYN nodes relative to the graph mean.
\textit{Graph Density} (GD) is the overall density of the semantic
edge set.
\textit{Revision Depth} (RvD) is the mean positional gap between
the endpoints of backtracking edges.

\paragraph{Metacognition.}
\textit{Critique/Hypothesis Ratio} (CHR) measures how often the model
critiques its own hypotheses, as the ratio of CRT to HYP nodes.
\textit{Hedging Density} (HD) is the fraction of TUs containing
explicit uncertainty markers (e.g.\ \textit{might}, \textit{perhaps},
\textit{possibly}, \textit{likely}, \textit{unclear}).
\textit{Perspective Taking} (PT) is the fraction of RFR nodes,
measuring how consistently the model reframes the problem throughout
the trace.

\paragraph{Efficiency.}
\textit{Token/Idea} (TPI) measures verbal economy as total trace
tokens divided by UPC.
\textit{Redundancy} (RR) is the fraction of TU pairs with cosine
similarity above 0.75.
\textit{Avg Tokens} (AT) and \textit{Avg TUs} (ATU) are the mean
token count and TU count per trace, providing verbosity and
granularity baselines.

\paragraph{5D-CP construction.}
Each metric is z-scored globally across all traces in the study corpus.
The 5D-CP for a single trace is:
\begin{equation}
\begin{split}
  \mathrm{CP}_d &= \frac{1}{|\mathcal{M}_d|}
    \sum_{m \in \mathcal{M}_d} z_m, \\[2pt]
  &d \in \{\text{\small Breadth, Depth, Structure,} \\
  &\qquad\;\; \text{\small Metacog., Efficiency}\}
\end{split}
\label{eq:5dcp}
\end{equation}

where $\mathcal{M}_d$ is the set of active metrics in dimension $d$
and $z_m$ is the global z-score of metric $m$ for that trace.
A model-level 5D-CP is the mean over all its traces.

\section{Experimental Protocol}
\label{sec:protocol}
We evaluate seven native reasoning models on 200 open-ended questions
across ten cognitive domains, collecting \textbf{three} independent runs per
model--question pair for a total of 4{,}200 reasoning traces.
All models produce extended chain-of-thought inside \texttt{<think>}
tags prior to their final answer.
Unlike supervised output, this pre-answer thinking is trained via
outcome-based reinforcement rather than token-level imitation, so its
internal structure reflects emergent reasoning behaviour rather than
learned surface formatting, making it a principled locus for
structural analysis.

The selection is constrained by compute budget and collection 
time; larger or additional native reasoning models 
(e.g.\ DeepSeek-R1, QwQ-32B) are left for future work.

\subsection{Models and Infrastructure}

We include seven publicly available native reasoning models spanning a
wide range of scales and architectures (Table~\ref{tab:models}).
Two models use sparse mixture-of-experts (MoE) routing
(Qwen3.5-35B-A3B and Nemotron-Super-120B-A12B), with active
parameter counts substantially smaller than their total footprint.
Traces are collected via a unified OpenAI-compatible endpoint
(vLLM-served for locally hosted models); the \texttt{<think>} block
is extracted verbatim and stored as the unit of analysis. \footnote{All open-weight model traces, extracted Thought Graphs, 
metric vectors, the question dataset, cue-phrase lexicons, 
and pipeline code will be released at publication.}

\begin{table}[t]
\centering\small
\begin{tabular}{p{3.0cm}lcc}
\toprule
\textbf{Model} & \textbf{Provider} & \textbf{Total} & \textbf{Active} \\
\midrule
GLM-4.7-Flash \cite{Zeng2025GLM45AR}           & ZhipuAI    &  30B  &  3B  \\
Phi-4-reasoning  \cite{Phi4}        & Microsoft  & 14B  & 14B  \\
Gemma-4-31B   \cite{Gemma-4-31B-IT}           & Google     & 31B  & 31B  \\
Qwen3.5-35B-A3B   \cite{qwen3.5}       & Alibaba    & 35B  &  3B  \\
Mistral-Medium-3.5  \cite{Mistral-Medium-3.5-128B}     & Mistral AI & 128B & 128B \\
Nemotron-Super-120B-A12B \cite{nvidia_nemotron_3_2025} & NVIDIA     & 120B & 12B  \\
GPT-OSS-120B   \cite{gptoss}         & OpenAI     & 117B & 5.1B \\
\bottomrule
\end{tabular}
\caption{Models evaluated in ThinkProbe. Active parameters reflect
  per-token compute for MoE models.}
\label{tab:models}
\vspace{-5mm}
\end{table}

\subsection{Question Dataset}

ThinkProbe uses a curated set of 200 open-ended questions distributed
evenly across ten cognitive domains (20 questions each,
Table~\ref{tab:domains}).
Questions were manually authored and LLM-assisted, then reviewed by
expert annotators who verified that every question is genuinely
open-ended, admitting no single correct answer and requiring the
model to weigh competing considerations rather than retrieve a fact.
All 200 questions passed this review.

Open-ended questions are essential to structural analysis: on
closed problems, answer correctness dominates and structural
variation is a secondary confound.
By removing the correctness signal entirely, we ensure that the
reasoning trace is the object of evaluation, not a by-product of it.

The ten domains span normative, empirical, strategic, and creative
reasoning, capturing the breadth of cognitive demands placed on
frontier models in practice.

\begin{table}[t]
\centering\small
\begin{tabular}{lp{4.0cm}}
\toprule
\textbf{Domain} & \textbf{Characterisation} \\
\midrule
Ethical Dilemmas          & Moral conflicts requiring value prioritisation under uncertainty \\
Policy Design             & Institutional design and regulatory trade-offs \\
Strategic Planning        & Multi-step planning under resource and goal constraints \\
Scientific Speculation    & Hypothesis generation and causal reasoning under uncertainty \\
Creative Problem Solving  & Generative ideation with novel or hard constraints \\
Interpersonal Reasoning   & Social dynamics and relational decision-making \\
Economics \& Markets      & Market dynamics, incentive structures, and economic trade-offs \\
Geopolitics               & International relations, power dynamics, and statecraft \\
Environmental Ethics      & Ecological value judgements and sustainability dilemmas \\
Philosophy \& Metaphysics & Abstract conceptual and ontological inquiry \\
\bottomrule
\end{tabular}
\caption{Ten cognitive domains in ThinkProbe (20 questions each).}
\label{tab:domains}
\vspace{-5mm}
\end{table}

\subsection{Collection Setup}

Each model is queried independently on every question for three runs,
sampling with non-zero temperature to capture run-to-run stochasticity
(hyperparameters reported in Appendix~\ref{app:collection}).
The \texttt{<think>} block is extracted verbatim from each response;
the final answer is discarded.
Traces exhibiting degenerate looping behaviour detected via repeated
normalised line hashing are retried up to three times; if looping
persists, the trace is truncated at the first repeated segment.
The full collection yields \textbf{4,200 traces}
(7 models $\times$ 200 questions $\times$ 3 runs).

\section{Results and Discussion}
\label{sec:results}
Across 4{,}200 reasoning traces, ThinkProbe reveals that models occupy
distinct, interpretable regions of the 5D cognitive space: no two
models share a profile, and the differences are large, reproducible,
and statistically significant across all 19 active metrics.
The following subsections characterise the profiles (§\ref{sec:profiles}),
quantify per-metric discriminability (§\ref{sec:discrim}), show that
domain identity is a secondary driver of reasoning structure
(§\ref{sec:domain}), and validate that each pipeline layer contributes
non-redundant signal (§\ref{sec:ablation}).

\subsection{Cognitive Profile Heterogeneity}
\label{sec:profiles}

\begin{figure}[t]
  \centering
  \includegraphics[width=\columnwidth]{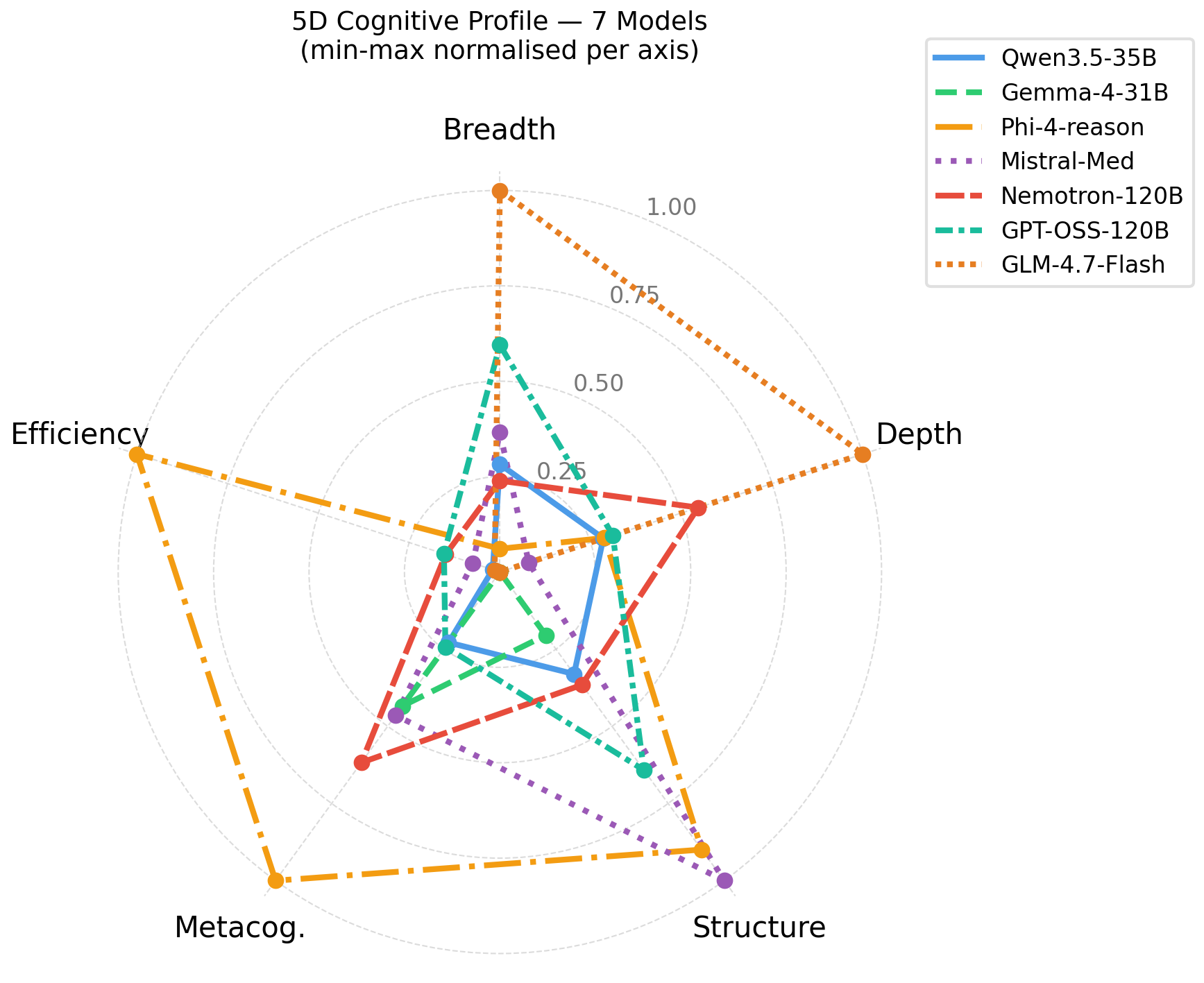}
  \caption{5D Cognitive Radar: per-model mean profiles (min--max
    normalised per axis). Each polygon represents one model's
    aggregate reasoning style across Breadth, Depth, Structure,
    Metacognitive, and Efficiency dimensions.}
  \label{fig:radar}
  \vspace{-5mm}
\end{figure}

Figure~\ref{fig:radar} and Table~\ref{tab:5d} show mean 5D-CP z-scores
for all seven models.
The profiles reveal three distinct behavioural regimes.

\textbf{High-efficiency, high-structure thinkers.}
Phi-4-reasoning stands alone on the Efficiency axis ($z{=}{+}1.27$),
accompanied by above-average Depth, Structure, and Metacognitive
scores.
It generates the most token-dense, structurally complex traces in the
corpus, but does so at the cost of Breadth ($z{=}{-}0.30$):
its reasoning drills deep rather than ranges wide.

\textbf{Broad-and-deep generalists.}
GLM-4.7-Flash leads on both Breadth ($z{=}{+}0.66$) and Depth
($z{=}{+}0.59$), with near-zero Structure and below-average Efficiency
and Metacognitive scores.
GPT-OSS-120B follows a similar breadth-forward pattern
($z{=}{+}0.28$) with a flatter profile elsewhere.

\textbf{Compact, low-variance models.}
Gemma-4-31B records the most negative scores on Breadth ($-$0.38),
Depth ($-$0.49), and Efficiency ($-$0.34), producing the shortest,
structurally simplest traces.
Qwen3.5-35B-A3B, Mistral-Med.-3.5, and Nemotron-120B cluster near the origin,
with moderate, undifferentiated profiles.

The geometric distance between extremes is large: the cosine similarity
between Phi-4-reasoning and Qwen3.5-35B-A3B is $-$0.91, confirming
near-orthogonal cognitive strategies despite similar task inputs.
Qwen3.5-35B-A3B and Gemma-4-31B are the most similar pair ($+$0.77).

\begin{table}[t]
\centering\scriptsize
\setlength{\tabcolsep}{3.5pt}
\resizebox{\columnwidth}{!}{%
\begin{tabular}{lccccc}
\toprule
\textbf{Model} & \textbf{Brdth} & \textbf{Depth} & \textbf{Struct} & \textbf{Metacog} & \textbf{Effic} \\
\midrule
Phi-4-reasoning          & $-$0.30 & $+$0.31 & $+$0.19 & $+$0.29 & $\mathbf{+1.27}$ \\
GLM-4.7-Flash            & $\mathbf{+0.66}$ & $\mathbf{+0.59}$ & $-$0.13 & $-$0.25 & $-$0.31 \\
GPT-OSS-120B             & $+$0.28 & $-$0.05 & $+$0.05 & $-$0.12 & $-$0.09 \\
Nemotron-120B            & $-$0.18 & $+$0.08 & $-$0.03 & $+$0.14 & $-$0.10 \\
Mistral-Med.-3.5              & $-$0.03 & $-$0.33 & $\mathbf{+0.27}$ & $+$0.01 & $-$0.22 \\
Qwen3.5-35B-A3B              & $-$0.06 & $-$0.12 & $-$0.10 & $-$0.10 & $-$0.31 \\
Gemma-4-31B              & $-$0.38 & $-$0.49 & $-$0.28 & $\mathbf{+0.00}$ & $-$0.34 \\
\midrule
KW $\varepsilon^2$       & 0.317   & 0.209   & 0.121   & 0.120& 0.385   \\
\bottomrule
\end{tabular}}
\caption{Mean 5D-CP z-scores per model (global normalisation across all traces). Bold marks the per-column maximum. Bottom row: KW $\varepsilon^2$ per dimension (all $p < 0.001$). Figure~\ref{fig:radar} visualises the same scores after min-max normalisation per axis to aid geometric comparison across dimensions.}
\label{tab:5d}
\vspace{-5mm}
\end{table}

\subsection{Metric Discriminability}
\label{sec:discrim}

\begin{figure}[t]
  \centering
  \includegraphics[width=\columnwidth]{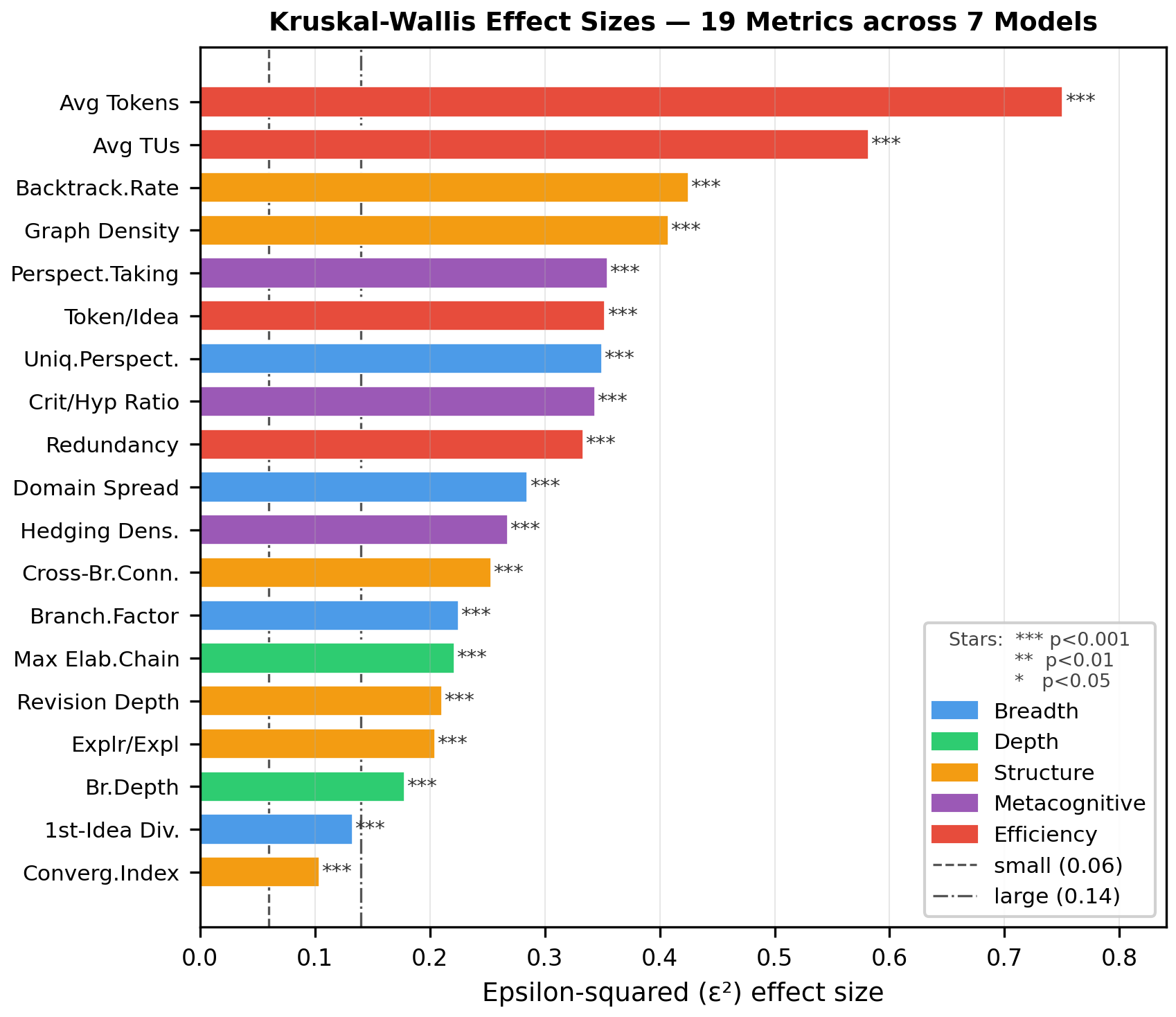}
  \caption{Kruskal--Wallis $\varepsilon^2$ effect sizes for all 19
    active metrics, sorted descending and colour-coded by 5D family.
    Dashed lines mark small ($\varepsilon^2=0.06$) and large
    ($\varepsilon^2=0.14$) effect thresholds. All metrics
    $p < 0.001$.}
  \label{fig:kw}
  \vspace{-5mm}
\end{figure}

Figure~\ref{fig:kw} reports Kruskal--Wallis $\varepsilon^2$ effect
sizes for all 19 active metrics.
Every metric reaches statistical significance ($p < 0.001$, 
uncorrected; all 19 survive Bonferroni correction for 19 
comparisons, $p < 0.000053$ threshold),
and effect sizes span a wide range: from $\varepsilon^2 = 0.75$
(Avg.\ Tokens) down to $\varepsilon^2 = 0.10$
(First-Idea Diversity).

The three largest effects Avg.\ Tokens ($0.75$), Avg.\ TUs ($0.58$),
and Backtracking Rate ($0.43$) are verbosity or structural-revision
proxies.
Notably, pure structural graph metrics that are length-independent
also achieve large effects: Perspective Taking ($0.35$),
Critique-to-Hypothesis Ratio ($0.35$), and Unique Perspectives ($0.35$)
all discriminate models comparably to raw verbosity measures.

At the lower end, Cross-Branch Connectivity ($0.13$), Convergence
Index ($0.12$), Mean Branch Depth ($0.10$), and First-Idea Diversity
($0.10$) register medium effects small in absolute magnitude but
consistent across 4{,}200 traces, confirming that even fine-grained
graph-structural differences are reproducibly model-specific rather
than noise.

\subsection{Model Identity Dominates Domain}
\label{sec:domain}

To determine whether observed profile differences reflect genuine
model behaviour or are simply driven by question content, we repeat
the Kruskal--Wallis analysis with \emph{domain} as the grouping
variable.
The contrast is stark: the top model-level effect size is
$\varepsilon^2 = 0.75$ (Avg.\ Tokens), while the highest domain-level
effect across all 19 active cognitive metrics is
$\varepsilon^2 = 0.19$ (Hedging Density) a fourfold gap.
For structural graph metrics such as Branching Factor and
Graph Density, domain effects are near-zero
($\varepsilon^2 < 0.02$), whereas model-level effects for the same
metrics exceed $0.22$.

This asymmetry holds across all five 5D dimensions
(Figure~\ref{fig:domain_heatmap}, supplementary): model rows in the
profile heatmap show strong, consistent colouring while domain columns
are near-uniform.
We conclude that reasoning structure in these traces is a stable,
model-level property: the same model reasons similarly regardless of
whether it is asked about geopolitics, ethics, or economics.
Domain identity shifts surface content but leaves the underlying
cognitive architecture largely intact.

The single exception is the Structure dimension, where domain effects ($\varepsilon^2$=0.156) marginally exceed model effects ($\varepsilon^2$=0.121), reflecting genuine variation in graph topology across question types — particularly cross-branch connectivity and convergence behaviour, which are sensitive to whether a question demands integration across perspectives or sustained single-thread elaboration.

Prompt design is a potential confound: an explicit elicitation prompt inflates UPC 2$\times$ on Qwen3.5-35B-A3B (Appendix~\ref{app:collection}). However, the empty, pure, and normal conditions — spanning naturally occurring deployment prompts — produce near-identical profiles (BF: 0.42/0.43/0.43, UPC: 5.6/5.2/4.7), and prompt condition is held constant across all seven models, ruling out prompt variation as a driver of between-model differences.

\subsection{Pipeline Layer Ablation}
\label{sec:ablation}

Table~\ref{tab:ablation} reports mean metric values under four
cumulative pipeline configurations on a held-out set of 50 traces.
The results confirm that each layer contributes non-redundant signal.

With structural boundaries only, all typed-edge metrics are zero by
construction: no branching, no backtracking, no cross-branch links.
Only metrics derivable from sequential node order
(Exploration/Exploitation Ratio, Critique-to-Hypothesis Ratio)
are non-zero at this stage.

Adding the semantic trajectory layer (L3) activates the branching
family: Branching Factor rises to 0.22, Max Elaboration Chain to 3.18,
and Graph Density to 0.17.
This layer is responsible for the bulk of structural differentiation
between models.

The cross-segment layer (L4) uniquely activates Revision Depth
(0 $\to$ 3.87), which requires identifying backward-pointing arcs
spanning non-adjacent segments.
Backtracking Rate and Cross-Branch Connectivity also initialise at
this stage, though at low values.

The full pipeline produces the largest gain in Cross-Branch
Connectivity (0.05 $\to$ 0.22), reflecting the final resolution of
long-range semantic links across the complete graph.
Graph Density reaches its maximum (0.30).
No metric collapses to  zero across ablation stages, confirming that each layer adds structure without corrupting prior signal.

\begin{table}[t]
\centering\scriptsize
\setlength{\tabcolsep}{4pt}
\resizebox{\columnwidth}{!}{%
\begin{tabular}{lcccc}
\toprule
\textbf{Metric} & \textbf{Struct.} & \textbf{+L3} & \textbf{+L4} & \textbf{Full} \\
\midrule
Branching Factor        & 0.00 & 0.22 & 0.22 & 0.22 \\
Max Elab.\ Chain        & 0.00 & 3.18 & 3.18 & 3.18 \\
Graph Density           & 0.00 & 0.17 & 0.19 & \textbf{0.30} \\
Revision Depth          & 0.00 & 0.00 & \textbf{3.87} & 3.87 \\
Cross-Branch Conn.      & 0.00 & 0.00 & 0.05 & \textbf{0.22} \\
Backtracking Rate       & 0.00 & 0.00 & 0.04 & 0.02 \\
Convergence Index       & 0.00 & 0.10 & 0.07 & 0.09 \\
\bottomrule
\end{tabular}}
\caption{Mean metric values under cumulative pipeline ablation
  ($N{=}50$ traces). \textbf{Bold} marks the stage at which a metric
  first reaches its maximum. Struct.\ = structural boundaries only;
  +L3 = with semantic trajectory; +L4 = with cross-segment analysis;
  Full = complete pipeline.}
\label{tab:ablation}
\vspace{-7mm}
\end{table}

\section{Conclusion}
\label{section:conclusion}
\vspace{-3mm}
ThinkProbe extracts Thought Graphs and a 19-metric 5D cognitive
profile from LLM reasoning traces via a fully non-generative pipeline,
eliminating LLM-analysing-LLM circularity. Across 4{,}200 traces,
all 19 metrics significantly discriminate models, and between-model
variance exceeds between-domain variance across four of five
dimensions, establishing reasoning structure as a predominantly
stable model-level property invisible to accuracy-based evaluation,
with direct implications for model selection, deployment monitoring,
and the evaluation of tasks where ground truth does not exist.
Future work includes correlating 5D profiles with downstream task
performance to enable real-time structural confidence signals during
inference, expanding to instruct models to probe the boundary between
elicited and native chain-of-thought, and applying ThinkProbe to
open-ended deployment contexts(moral reasoning, creative ideation,
strategic planning), where accuracy-based evaluation is unavailable.


\section{Limitations}
\label{sec:limitations}

\paragraph{Rule-based node classifier.}
The current node classifier is fully deterministic, mapping boundary
classes to node types via fixed rules. This makes it transparent and
free of analyser-model bias, but it inherits the precision limits of
surface cue-phrase matching: our human study (Appendix~\ref{app:segval})
shows that contrastive connectives such as \textit{however} and
\textit{but} are occasionally over-assigned to the BACKTRACK class.
This does not alter model rankings (backtracking rate remains one of
the most discriminative metrics at $\varepsilon^2=0.43$), but future
work could replace the rule set with a fine-tuned discriminative
classifier to improve boundary sensitivity and capture node-type
distinctions not reliably detectable from surface cues alone.

\paragraph{Single embedding model.}
The extraction pipeline — soft-boundary detection, edge typing, and
all similarity thresholds — relies on a single sentence encoder
(\texttt{all-MiniLM-L6-v2}). Because every threshold is a per-trace
percentile of that trace's own similarity distribution, our design
is invariant to monotone rescaling of the embedding space, and we
therefore expect the rank-order discriminability findings to be robust
to encoder choice (§\ref{sec:method:pipeline}). This is an argument
from construction, however, not an empirical result: we have not
re-run the pipeline under an alternative encoder, and quantifying
sensitivity to embedding-model choice remains an open item.

\paragraph{Length confound.}
Three metrics are correlated with trace verbosity: Avg.\ Tokens
($\varepsilon^2=0.75$, the single largest effect in the corpus),
Avg.\ TUs ($\rho\approx0.90$ with trace length), and Token/Idea
($\rho\approx0.60$). They should be interpreted with caution when
comparing models with systematically different trace lengths.
We bound this effect directly in Appendix~\ref{app:length_robustness}:
removing all three length-correlated metrics reduces total profile
$\varepsilon^2$ by at most 14\% (Efficiency only) and leaves the other
four dimensions entirely unchanged, and the model-over-domain
dominance ratio is preserved throughout. The 5D-CP's discriminative
power is therefore structurally grounded rather than a verbosity
artefact.

\paragraph{Equal-weight aggregation.}
The 5D-CP aggregates metrics within each dimension by equal-weight
averaging. For Breadth, Depth, Metacognitive, and Efficiency, model
rankings are invariant to PCA-weighted aggregation (rank
$\rho \geq 0.893$). The Structure dimension contains a genuine
connectivity cluster (CBC, CI, GD mutually correlated at
$\rho = 0.57$--$0.78$), and its rankings shift under PCA weighting
(rank $\rho = 0.357$). Structure-dimension comparisons should be
interpreted with this sensitivity in mind.

\paragraph{Limited-scale human validation.}
Thought Graphs are validated via pipeline ablation, statistical
discriminability, and a 50-trace human annotation study
(Appendix~\ref{app:segval}), in which the pipeline matches or exceeds
inter-annotator agreement on major cognitive transitions. This
establishes a human validity baseline, but at modest scale and with
two annotators; whether the extracted graphs align with expert
judgement across a larger trace sample and a broader annotator pool
remains an open question.

\paragraph{Native reasoning models only.}
The framework requires \texttt{<think>} traces and was evaluated
exclusively on native reasoning models. Generalization to instruct
models with prompted chain-of-thought has not been tested and would
require re-examining the open-ended question assumption.

\paragraph{Trace faithfulness.}
The \texttt{<think>} content may not faithfully represent the model's
internal computation; recent work shows that reasoning traces can
involve post-hoc rationalisation~\cite{liu2026diagnosing}.
ThinkProbe characterises the \emph{expressed} reasoning structure,
not necessarily the underlying mechanism.

\paragraph{External validity.}
ThinkProbe establishes that structural profiles are stable and
model-discriminating, but does not correlate 5D dimensions with
human-judged qualities such as coherence, persuasiveness, or novelty.
This is partly by design (the framework targets settings where such
signals are unavailable), but a correlation study on a subset of
questions would substantially strengthen the practical claims and is 
the highest-priority next step.


\bibliography{custom}

\clearpage

\appendix

\twocolumn[\centering{\Large\bfseries Supplementary Materials for the paper: \\ ThinkProbe: Beyond Accuracy - Structural Profiling of Open-Ended LLM Reasoning Traces via Non-Generative Thought Graphs}\vspace{2em}]

\section{Collection Hyperparameters}
\label{app:collection}

\begin{table}[h]
\centering\small
\begin{tabular}{ll}
\toprule
\textbf{Parameter} & \textbf{Value} \\
\midrule
Temperature           & 0.7 \\
Max tokens            & 20{,}000 \\
Prompt variant        & \texttt{empty} (no system prompt) \\
Runs per question     & 3 \\
Loop detection        & repeated normalised-line hashing \\
Max retries (loop)    & 3 \\
Truncation fallback   & first repeated segment \\
Concurrency           & asyncio semaphore, 5 parallel requests \\
\bottomrule
\end{tabular}
\caption{Collection hyperparameters used for all 7 models in the
  main study.}
\label{tab:hyperparams}
\end{table}

\paragraph{Prompt variant study.}
We evaluated four system prompt conditions on Qwen3.5-35B-A3B
across 30 traces each (Table~\ref{tab:prompt_variants}).
The \texttt{empty} condition (no system prompt) was selected for
the main study to avoid interfering with the models' native
reasoning behaviour.

\begin{table}[h]
\centering\small
\begin{tabular}{lp{5.5cm}}
\toprule
\textbf{Variant} & \textbf{System prompt} \\
\midrule
\texttt{empty}    & (none) \\
\texttt{pure}     & ``You are a helpful assistant.'' \\
\texttt{normal}   & ``You are a helpful assistant. Think carefully
                    before answering.'' \\
\texttt{eliciting} & Detailed 7-step cognitive elicitation
                    (spread wide, reframe, go deep, challenge,
                    connect, reflect, converge) \\
\bottomrule
\end{tabular}
\caption{System prompt variants evaluated.}
\label{tab:prompt_variants}
\end{table}

\begin{table}[h]
\centering\small
\setlength{\tabcolsep}{4pt}
\begin{tabular}{lcccccc}
\toprule
\textbf{Variant} & \textbf{Avg Tok} & \textbf{Avg TUs}
  & \textbf{BF} & \textbf{BR} & \textbf{UPC} & \textbf{GD} \\
\midrule
\texttt{empty}    & 763  & 22.8 & 0.42 & 0.036 & 5.6  & 0.165 \\
\texttt{pure}     & 693  & 19.9 & 0.43 & 0.038 & 5.2  & 0.182 \\
\texttt{normal}   & 685  & 20.1 & 0.43 & 0.034 & 4.7  & 0.177 \\
\texttt{eliciting}& 2622 & 117.7 & 0.35 & 0.032 & 11.4 & 0.064 \\
\bottomrule
\end{tabular}
\caption{Key metrics by prompt variant (Qwen3.5-35B-A3B, $N{=}30$
  traces). BF = Branching Factor, BR = Backtracking Rate,
  UPC = Unique Perspectives, GD = Graph Density.
  \texttt{empty}, \texttt{pure}, and \texttt{normal} produce
  near-identical structural profiles; \texttt{eliciting} inflates
  token count 3.4$\times$ while reducing graph density, as the
  larger node count dilutes per-pair connectivity.}
\label{tab:prompt_study}
\end{table}

\section{Node and Edge Taxonomy}
\label{app:taxonomy}

\subsection*{Node Types}

\begin{table}[H]
\centering\small
\setlength{\tabcolsep}{4pt}
\begin{tabular}{llp{2.8cm}p{3.0cm}}
\toprule
\textbf{Type} & \textbf{Family} & \textbf{Definition}
  & \textbf{Trigger condition} \\
\midrule
HYP & Exploration  & A new hypothesis, conjecture, or
  initial framing of the problem
  & First TU of a trace; or BRANCH boundary class \\
RFR & Exploration  & Restates the problem from a
  different conceptual starting point
  & Lexical reframing cues
  (\emph{alternatively}, \emph{from another angle}) \\
JUS & Elaboration  & Provides evidence, justification,
  or reasoning support for a prior claim
  & SUPPORT boundary class \\
SPC & Elaboration  & Narrows or specifies a prior
  idea with detail or constraint
  & ELABORATION boundary class; soft NONE boundaries \\
CRT & Evaluation   & Identifies a flaw, limitation,
  or counter-argument in prior reasoning
  & BACKTRACK boundary class \\
CMP & Evaluation   & Contrasts two ideas, framings,
  or positions against each other
  & CONTRAST boundary class \\
MET & Evaluation   & Reflects on the quality or
  direction of the reasoning process itself
  & META boundary class \\
SYN & Convergence  & Integrates multiple prior threads
  into a unified position or conclusion
  & CONVERGENCE boundary class \\
\bottomrule
\end{tabular}
\caption{Eight node types in ThinkProbe, organized into four cognitive families.}
\label{tab:node_types}
\end{table}

\subsection*{Edge Types}

\begin{table}[h]
\centering\small
\setlength{\tabcolsep}{4pt}
\begin{tabular}{lp{2.8cm}p{3.0cm}}
\toprule
\textbf{Type} & \textbf{Definition} & \textbf{Assignment condition} \\
\midrule
SEQ  & Default sequential flow between
  adjacent TUs
  & NONE or META boundary; any untyped
  adjacent pair \\
BRCH & A new reasoning branch diverges
  from the current thread
  & BRANCH boundary class (L3 semantic
  trajectory) \\
ELAB & The target TU elaborates or
  extends the source
  & ELABORATION boundary; confirmed by
  embedding similarity \\
BACK & The target TU revises or corrects
  a non-adjacent prior TU
  & BACKTRACK boundary spanning $> 1$
  segment (L4 cross-segment) \\
SYNT & The target TU synthesises content
  from multiple source TUs
  & CONVERGENCE boundary with
  multi-parent linking \\
CRIT & The target TU critiques the source
  & CRT node type at target with BACK-type
  arc to a non-adjacent source \\
\bottomrule
\end{tabular}
\caption{Six edge types in ThinkProbe Thought Graphs.
  SEQ edges represent the default sequential reading order.
  Typed edges (SEQ, BRCH, ELAB, BACK, SYNT, CRIT)
  encode semantic relationships identified by the extraction
  pipeline.}
\label{tab:edge_types}
\end{table}

\section{Metric Definitions}
\label{app:metrics}

Table~\ref{tab:metrics_expand} defines all 19 active metrics.
Let $G = (V, E)$ denote a Thought Graph; $E_X \subseteq E$ the subset
of edges of type $X$; $E_{\text{sem}} = E \setminus E_{\text{SEQ}}$
the set of all non-sequential (semantic) edges; $\text{pos}(u)$ the
sequential index of node $u$; and $d_{\text{in}}(v)$ the in-degree
of node $v$.

\begin{table*}[t]
\centering\small
\setlength{\tabcolsep}{4pt}
\begin{tabular}{llp{7cm}p{4.6cm}}
\toprule
\textbf{Metric} & \textbf{Dim.} & \textbf{Formula} &
\textbf{Interpretation} \\
\midrule
Branching Factor (BF)
  & Breadth
  & $|E_{\text{BRCH}}| \;/\; \max(|V|,1)$
  & Rate of new reasoning branches per node \\
Unique Perspectives (UPC)
  & Breadth
  & $|\{v \in V : \text{type}(v)=\text{RFR}\}|$
  & Count of explicit reframings of the problem \\
Domain Spread (DS)
  & Breadth
  & $\bigl|\text{clusters}\bigl(\{e_v : v \in V_{\text{BRS}}
    \cup V_{\text{HYP}}\},\;\delta{=}0.45\bigr)\bigr|$
  & Semantic cluster count of hypothesis-space
    embeddings \\
First-Idea Diversity (FID)
  & Breadth
  & $\tfrac{1}{\binom{k}{2}}
    \displaystyle\sum_{i<j}(1-\cos(e_i,e_j))$,\;
    $k{=}\min(3,|V_{\text{HYP}}|)$
  & Mean pairwise cosine distance of the
    first three HYP node embeddings \\
\midrule
Max Elaboration Chain (MEC)
  & Depth
  & $\text{longest-path}\bigl(V,\,E_{\text{ELAB}}\bigr)$
  & Maximum depth of elaborative reasoning \\
Mean Branch Depth (MBD)
  & Depth
  & $\overline{d_{\text{sem}}(v)}$,\;
    $v$ non-root in $G_{\text{sem}}=(V,\,E{\setminus}E_{\text{SEQ}})$
  & Mean shortest-path depth from roots in
    the full semantic subgraph \\
\midrule
Expl./Exploit.\ Ratio (EER)
  & Structure
  & $|V_{\text{EXPL}}| \;/\;
    \max(|V_{\text{ELAB}}|,\,1)$
  & Balance between exploration and
    elaboration nodes \\
Backtracking Rate (BR)
  & Structure
  & $|E_{\text{BACK}}| \;/\; \max(|E_{\text{sem}}|, 1)$
  & Fraction of semantic edges that revise
    prior reasoning \\
Cross-Branch Conn.\ (CBC)
  & Structure
  & $\displaystyle\frac{
      |\{(B_i,B_j):i{<}j,\;\exists\,e{\in}
        E_{\text{SYNT}}
        \text{ crossing }B_i,B_j\}|}
      {\binom{|\mathcal{B}|}{2}}$
  & Fraction of branch-component pairs
    linked by a synthesis edge \\
Convergence Index (CI)
  & Structure
  & $\displaystyle\sum_{v\in V_{\text{SYN}}}
    \!d_{\text{in}}(v)
    \;\Big/\;
    \bigl(|V|\cdot\overline{d_{\text{in}}}\bigr)$
  & Degree to which synthesis nodes
    aggregate multiple threads \\
Graph Density (GD)
  & Structure
  & $|E_{\text{sem}}| \;/\;
    \bigl(|V|(|V|-1)\bigr)$
  & Overall semantic connectivity of
    the graph \\
Revision Depth (RvD)
  & Structure
  & $\displaystyle\operatorname{mean}_{
      (u,v)\in E_{\text{BACK}}}
    |\text{pos}(u)-\text{pos}(v)|$
  & Average positional span of backtracking
    arcs \\
\midrule
Critique/Hyp.\ Ratio (CHR)
  & Metacog.
  & $|V_{\text{CRT}}| \;/\; \max(|V_{\text{HYP}}|, 1)$
  & Ratio of self-critiques to hypotheses
    generated \\
Hedging Density (HD)
  & Metacog.
  & $|\{v{\in}V : v \text{ matches} \geq 1
    \text{ hedge pattern}\}| \;/\; |V|$
  & Fraction of nodes containing epistemic
    uncertainty markers \\
Perspective Taking (PT)
  & Metacog.
  & $|V_{\text{RFR}}| \;/\; \max(|V|, 1)$
  & Fraction of nodes that reframe the
    problem from a new angle \\
\midrule
Token per Idea (TPI)
  & Efficiency
  & $\text{tokens} \;/\; |V_{\text{RFR}}|$;\;
    falls back to $\text{tokens} \;/\; |V|$
    if $|V_{\text{RFR}}|{=}0$
  & Mean tokens expended per unique
    perspective \\
Redundancy Ratio (RR)
  & Efficiency
  & $|\{(i,j):i{<}j,\;\cos(e_i,e_j){>}0.75\}|
    \;/\; \tbinom{|V|}{2}$
  & Fraction of node pairs with near-duplicate
    content \\
\midrule
Avg.\ Tokens
  & Summary
  & Mean token count per trace
  & Overall trace verbosity \\
Avg.\ TUs
  & Summary
  & Mean thought-unit count per trace
  & Overall trace segmentation density \\
\bottomrule
\end{tabular}
\caption{All 19 active metrics.}
\label{tab:metrics_expand}
\end{table*}

\section{Extraction Pipeline}
\label{app:pipeline}
The pipeline transforms a raw \texttt{<think>} trace into a Thought
Graph through four sequential, non-generative procedures.
\textsc{StructuralSegmentation} identifies hard boundaries using a
compiled cue-phrase lexicon (e.g.\ \emph{wait}, \emph{actually},
\emph{on the other hand}) and assigns a semantic \texttt{BoundaryClass}
to each transition; short spans are merged to enforce a minimum
granularity of three sentences or 50 tokens.
\textsc{TextTiling} complements hard boundaries with soft ones:
sentence embeddings from \textsc{all-MiniLM-L6-v2} are compared in
sliding blocks, and positions whose cosine similarity falls below the
trace-local 30th-percentile threshold are inserted as \textsc{None}
boundaries.
\textsc{SemanticTrajectory} assigns a typed edge to every adjacent TU
pair within a window of four, promoting pairs with high embedding
similarity to \textsc{Elab} and pairs with large semantic shift to
\textsc{Brch}.
\textsc{CrossSegmentAnalysis} then resolves long-range structure: it
batch-computes all pairwise TU similarities in a single call and
adds \textsc{Back} arcs for backward-pointing CRT/MET nodes,
\textsc{Synt} arcs where SYN nodes integrate multiple threads.
\textsc{GraphAssembly} materialises the Thought Graph and validates
three invariants: no self-loops, the \textsc{Elab}-only subgraph is
acyclic, and every node is reachable from $v_0$.
No generative LLM is involved at any stage.

\begin{algorithm*}[t]
\caption{ThinkProbe Non-Generative Extraction Pipeline}
\label{alg:pipeline}
\begin{algorithmic}[1]
\Require CoT trace $T$ (raw \texttt{<think>} text)
\Ensure Thought Graph $G = (V, E, \lambda_V, \lambda_E)$

\vspace{4pt}
\State \textbf{// Layer 1 — Structural Segmentation}
\State Split $T$ into sentences; detect paragraph breaks
\State For each sentence boundary, match cue-phrase lexicon
\State Assign $\text{BoundaryClass} \in$ \{BACKTRACK, BRANCH, META,
\Statex \hspace{2.5em} CONVERGENCE, ELABORATION, CONTRAST, SUPPORT\}
\State Merge spans $< 3$ sentences or $< 50$ tokens with successor
\State Paragraph breaks and unmatched boundaries $\to$ NONE

\vspace{4pt}
\State \textbf{// Layer 2 — Soft Boundary Detection (TextTiling)}
\State Encode all sentences with \textsc{all-MiniLM-L6-v2}
\State Slide window of size 3; compute block cosine similarities
\State $\tau \gets$ 30th percentile of within-trace similarities
\State Insert NONE boundary at local minima below $\tau$
\Statex \hspace{2em} (prominence $\geq 0.15$ via \texttt{find\_peaks})

\vspace{4pt}
\State \textbf{// Layer 3 — Semantic Trajectory}
\State Embed each TU with \textsc{all-MiniLM-L6-v2}
\State For adjacent TU pairs within window $w = 4$:
\State \hspace{2em} Compute embedding cosine similarity $s$
\State \hspace{2em} If $s > \tau_{\text{elab}}$: promote edge to ELAB
\State \hspace{2em} If semantic shift $> \tau_{\text{brch}}$: assign BRCH edge
\State Build initial edge set $E_0$ with SEQ, ELAB, BRCH

\vspace{4pt}
\State \textbf{// Layer 4 — Cross-Segment Analysis}
\State For all TU pairs $(u, v)$ with $|\text{pos}(u) - \text{pos}(v)| > 1$:
\State \hspace{2em} If $\text{BoundaryClass}(v) = \text{BACKTRACK}$ and backward arc:
\Statex \hspace{3.5em} add CRIT edge $(v \to u)$
\State \hspace{2em} If $\text{BoundaryClass}(v) = \text{META}$ and backward arc:
\Statex \hspace{3.5em} add BACK edge $(v \to u)$
\State \hspace{2em} If $\text{BoundaryClass}(v) = \text{CONVERGENCE}$ and multiple predecessors:
\Statex \hspace{3.5em} add SYNT edges from source threads
\State Batch all pair embeddings; compute pairwise cosine
\vspace{4pt}
\State \textbf{// Graph Assembly}
\State For each TU $t_i$: create node $v_i$, assign NodeType via
\Statex \hspace{2em} $\text{BOUNDARY\_NODE\_MAP}[\text{BoundaryClass}(t_i)]$
\State $V \gets \{v_0, \ldots, v_{n-1}\}$,\quad
       $E \gets E_0 \cup E_{\text{BACK}} \cup E_{\text{SYNT}} \cup E_{\text{CRIT}}$
\State Validate: no self-loops; ELAB subgraph acyclic;
\Statex \hspace{2em} all nodes reachable from $v_0$
\State \Return $G = (V, E, \lambda_V, \lambda_E)$
\end{algorithmic}
\end{algorithm*}

\section{Question Dataset}
\label{app:dataset}

\subsection*{Construction Process}

Questions were constructed in two stages.
In the \emph{authoring stage}, seed questions were manually written by
the authors for each domain, then expanded with LLM assistance to
reach a target of 20 questions per domain while maintaining topical
diversity within the domain.
In the validation stage, two expert annotators independently 
reviewed all 200 questions against a single binary criterion: 
the question must be genuinely open-ended, admitting no single 
correct answer and requiring the model to weigh competing 
considerations rather than retrieve a fact. Any question on 
which the two annotators disagreed was rejected and replaced 
with a new candidate until unanimous agreement was reached. 
The final 200 questions represent items on which both annotators 
agreed without exception, confirming that the dataset contains 
no borderline or factual-retrieval questions.

\subsection*{Sample Questions}

Table~\ref{tab:sample_questions} provides one representative question
per domain.

\begin{table*}[t]
\centering\small
\setlength{\tabcolsep}{5pt}
\begin{tabular}{lp{11.5cm}}
\toprule
\textbf{Domain} & \textbf{Sample question (truncated)} \\
\midrule
Ethical Dilemmas
  & A self-driving car can swerve to avoid two jaywalking pedestrians
    or continue straight and hit one legally crossing pedestrian.
    How should the car be programmed, and who bears moral and legal
    responsibility? \\[4pt]
Policy Design
  & A government proposes replacing all welfare programs with a
    universal basic income of \$1{,}500/month per adult. Advocates
    cite elimination of poverty traps; critics warn of abandoning
    targeted support. Design and defend a coherent policy position. \\[4pt]
Strategic Planning
  & A startup with \$5M runway and 10\% MoM growth faces a
    well-funded competitor entering its niche. Should it defend the
    niche, pivot to a mass market, or seek acquisition? \\[4pt]
Scientific Speculation
  & Reproducible evidence emerges that human consciousness persists
    for several minutes after clinical brain death. What is the
    scientifically rigorous response, and how should medicine, law,
    and ethics adapt? \\[4pt]
Interpersonal Reasoning
  & A close friend reveals their partner is having an affair and
    asks you not to intervene. Your own partner is also friends with
    the couple. How do you navigate competing loyalties? \\[4pt]
Creative Problem Solving
  & Transform a car-dependent city of one million into one where
    private car ownership is illegal. Suburbs are highway-oriented;
    elderly residents fear losing independence. Propose a credible
    transition plan. \\[4pt]
Geopolitics
  & Was NATO's post-Cold War eastward expansion a legitimate exercise
    of sovereign choice, a strategic provocation, or both? How should
    this shape current security architecture? \\[4pt]
Economics \& Markets
  & Should central banks elevate employment to an equal or primary
    mandate alongside price stability? What are the consequences for
    monetary policy independence and distributional outcomes? \\[4pt]
Philosophy \& Metaphysics
  & If determinism is true, does it undermine moral responsibility?
    Can compatibilist accounts preserve the practices of praise,
    blame, and punishment? \\[4pt]
Environmental Ethics
  & Should nature be granted legal rights or standing independent of
    human interests? How would such rights be represented, enforced,
    and balanced against development needs? \\
\bottomrule
\end{tabular}
\caption{One representative question per domain (truncated for space).
  Full questions average 87 words. The complete dataset will be
  released at publication.}
\label{tab:sample_questions}
\end{table*}

\section{Directed Cycles in Thought Graphs}
\label{app:cycles}

\paragraph{Motivation.}
A central design choice in our framework is to represent reasoning traces as full directed graphs rather than trees or DAGs. Cycles are the formal signature of iterative refinement: a model revisits an earlier reasoning state after elaborating or critiquing it, then continues forward. To verify that this expressivity is empirically necessary — and not a theoretical nicety — we enumerate all simple directed cycles across the 4,200 traces in our main study.

\begin{figure}[h]
  \centering
  \includegraphics[width=\columnwidth]{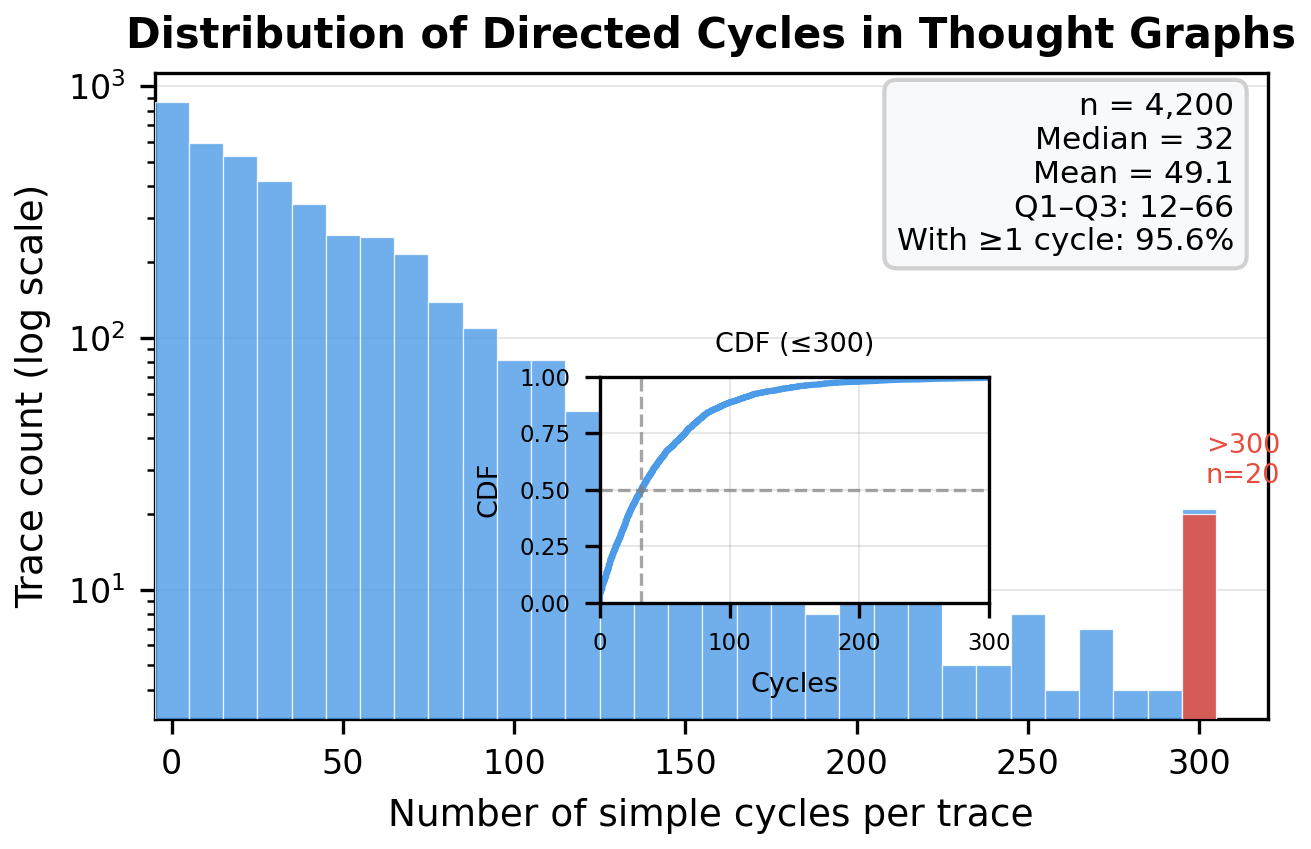}
  \caption{Distribution of simple directed cycle counts per trace (n=4,200), log-scale. Red bar = traces with >300 cycles. Inset: empirical CDF over the [0,300] range. Dashed crosshairs mark the median (32).}
  \label{fig:cycle_dist}
\end{figure}

\paragraph{Prevalence and distribution.}
Directed cycles appear in \textbf{95.6\%} of all traces (4,014 / 4,200), confirming that cyclic reasoning is the norm, not an edge case. The cycle count distribution is heavily right-skewed (Figure~\ref{fig:cycle_dist}): median 32, mean 49.1, Q1–Q3 of 13–66, and a long tail reaching 7,051 cycles in the most recursive trace. The 206,429 cycles collected across all traces have a median length of \textbf{14 nodes}, indicating that these are genuine multi-step iterative loops — not degenerate 2-node back-edges — and thus cannot be approximated by simple backtracking edges in a DAG.

\begin{figure}[h]
  \centering
  \includegraphics[width=\columnwidth]{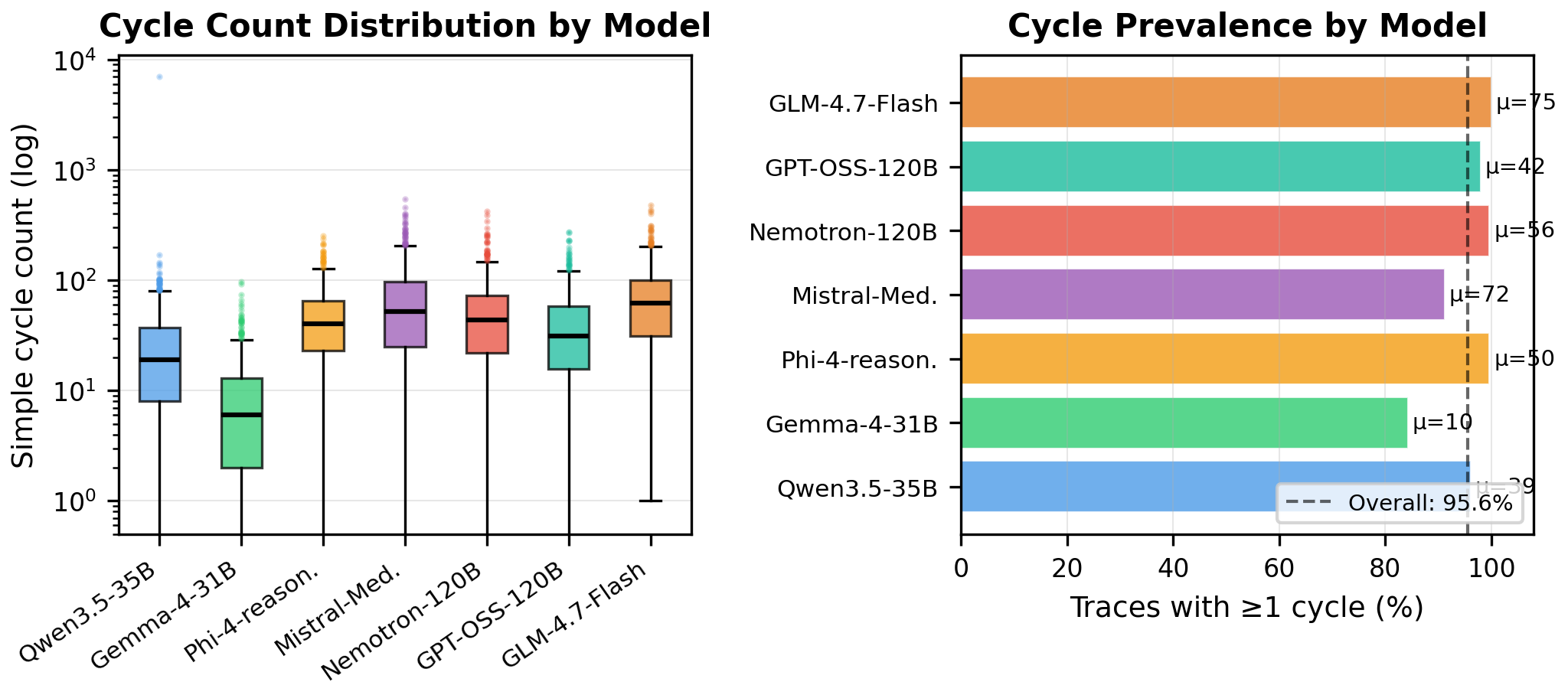}
  \caption{\textit{Left}: per-model cycle count distributions (log-scale box plots). \textit{Right}: fraction of traces containing at least one directed cycle; dashed line marks the overall rate (95.6\%). $\mu$ annotations show per-model means.}
  \label{fig:cycle_model}
\end{figure}

\paragraph{Per-model variation.}
Table~\ref{tab:cycle_model} and Figure~\ref{fig:cycle_model} show that cycle counts vary substantially across models, mirroring the Depth and Structure dimensions of the 5D-CP. GLM-4.7-Flash exhibits the highest cyclic activity ($\mu=75.4$, 100\% prevalence), consistent with its top Breadth and Depth profile scores. Gemma-4-31B is the clear outlier ($\mu=9.6$, 84.3\% prevalence) — markedly less recursive than all other models — which aligns with its uniformly low 5D-CP scores. Phi-4-reasoning and Nemotron-120B share near-identical prevalence (99.7\%) but differ in mean count (50 vs.\ 56), reflecting differences in trace length rather than reasoning style.

\begin{table}[h]
\centering
\small
\setlength{\tabcolsep}{5pt}
\begin{tabular}{lrrrrr}
\toprule
\textbf{Model} & \textbf{Mean} & \textbf{Median} & \textbf{Q1} & \textbf{Q3} & \textbf{\% $\geq$1} \\
\midrule
GLM-4.7-Flash       & 75.4 & 50 & 25 & 89  & 100.0 \\
Mistral-Med.-3.5    & 72.4 & 53 & 26 & 93  & 91.2  \\
Nemotron-120B       & 55.8 & 37 & 19 & 69  & 99.7  \\
Phi-4-reasoning     & 49.8 & 35 & 17 & 63  & 99.7  \\
GPT-OSS-120B        & 42.3 & 28 & 13 & 55  & 98.0  \\
Qwen3.5-35B-A3B         & 38.8 & 26 & 12 & 51  & 96.2  \\
Gemma-4-31B         &  9.6 &  6 &  2 & 13  & 84.3  \\
\midrule
\textbf{All}        & 49.2 & 32 & 13 & 66  & 95.6  \\
\bottomrule
\end{tabular}
\caption{Simple directed cycle statistics per model. Q1/Q3 are 25th/75th percentiles. Models sorted by mean.}
\label{tab:cycle_model}
\end{table}

\begin{figure}[h]
  \centering
  \includegraphics[width=\columnwidth]{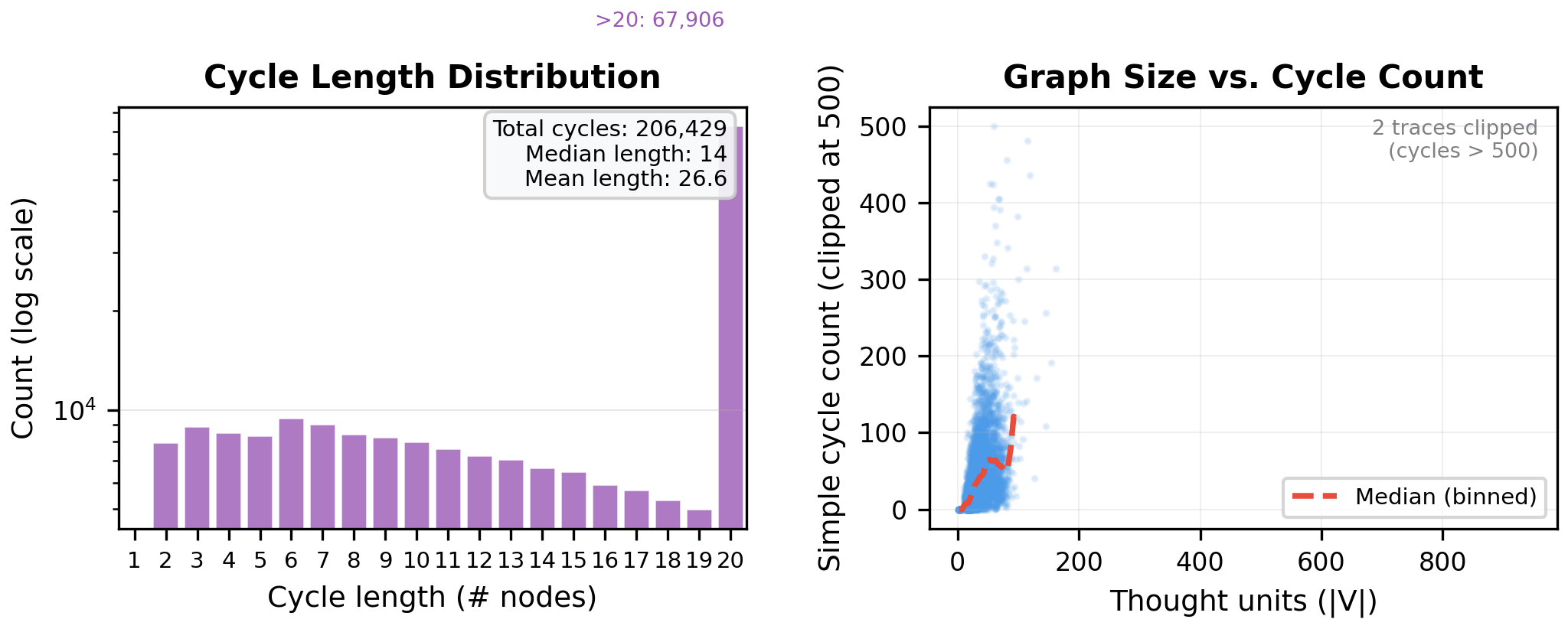}
  \caption{\textit{Left}: Distribution of simple cycle lengths (log-scale). The bar at position 20 aggregates all cycles of length $\geq 20$ (67,906 total). Median length = 14 nodes. \textit{Right}: scatter of graph size ($|V|$) vs.\ cycle count per trace; dashed line shows the binned median, illustrating super-linear growth.}
  \label{fig:cycle_length}
\end{figure}

\paragraph{Cycle length and graph size.}
Figure~\ref{fig:cycle_length} shows the cycle length distribution and its relationship to graph size. Short cycles (length 2–6) are most frequent but medium-length cycles (7–20 nodes) are nearly as common, reflecting the fact that iterative refinement often spans several reasoning steps before returning to a prior state. Cycle count grows super-linearly with the number of thought units $|V|$, as expected for dense directed graphs (Figure~\ref{fig:cycle_length}, right panel).

\paragraph{Implication.}
The near-universal presence of cycles (95.6\%) and their non-trivial length (median 14 nodes) provide direct empirical justification for the directed-graph representation. A DAG-constrained model — such as LCoT2Tree~\cite{jiang2025LCOTTree} — would need to either discard these structures entirely or approximate them with forward-only edges, losing the iterative refinement signal that our Depth and Structure metrics are designed to capture.

\paragraph{Cycle structure by edge-type subgraph.}
To assess whether cycles reflect genuine reasoning phenomena rather
than artefacts of dense edge assignment, we decomposed cycle
prevalence by edge-type subgraph across all 4,200 traces.
The \textsc{elab}+\textsc{back} subgraph produces zero cycles in
all traces, confirming that \textsc{elab} edges form a forward DAG
and \textsc{back} edges are strictly backward arcs.
Cycles arise exclusively from \textsc{synt} edges interacting with
the sequential backbone: the \textsc{seq}+\textsc{synt} subgraph
recovers 97.5\% cycle prevalence and a mean of 67.8 cycles per
trace, matching the full-graph rate (95.6\%, mean 49.2).
Each \textsc{synt} edge connects a synthesis node~$v_j$ back to
a representative of an earlier segment~$i < j$, creating a directed
cycle of length $(j - i + 1)$ through the \textsc{seq} backbone.
Cycles therefore reflect \emph{convergence} (later reasoning
states drawing simultaneously from multiple prior segments) not
iterative self-correction, which is captured independently by
\textsc{back}-edge metrics (backtracking rate, revision depth).

\section{Length Confound Robustness Check}
\label{app:length_robustness}

\paragraph{Motivation.}
The two highest-$\varepsilon^2$ metrics in the full Kruskal–Wallis sweep are
\textsc{avg\_tokens} ($\varepsilon^2=0.750$) and \textsc{avg\_tus} ($\varepsilon^2=0.581$),
both strongly correlated with raw trace length ($\rho\approx0.90$ and $\rho\approx0.60$
with token count, respectively).
A natural concern is whether the 5D-CP's cross-model discriminability is primarily
driven by verbosity differences already captured by simpler length benchmarks.
We address this directly by recomputing all five dimension scores under a
\emph{length-excluded} variant of the profile.

\paragraph{Excluded metrics.}
We remove three length-correlated metrics:
(i)~\textsc{avg\_tokens} and \textsc{avg\_tus}, which are direct length proxies, and
(ii)~\textsc{token\_per\_idea} (\textsc{TPI}), whose numerator is the raw token count
(normalised by $|V_\text{RFR}|$, but still correlated with trace length at $\varepsilon^2=0.351$).
The remaining 16 metrics are re-standardised and averaged within each dimension as before.
This is a conservative exclusion: it removes the only length-bearing component from
Efficiency and leaves the other four dimensions structurally unchanged.

\paragraph{Results.}
Table~\ref{tab:length_robustness} reports $\varepsilon^2$ for both model and domain
groupings under the full 5D-CP and the length-excluded variant.
Four of five dimensions — Breadth, Depth, Structure, and Metacognitive —
are completely unaffected ($\Delta\varepsilon^2=0.000$), because none of their
constituent metrics involve token counts.
Efficiency decreases from $\varepsilon^2=0.385$ to $0.332$ ($-14\%$), driven entirely
by the removal of TPI; the Efficiency metric (redundancy ratio, $\varepsilon^2=0.332$)
measures semantic overlap between thought units, not trace length.

\begin{table}[h]
\centering
\small
\setlength{\tabcolsep}{5pt}
\begin{tabular}{lcccc}
\toprule
 & \multicolumn{2}{c}{\textbf{Full 5D-CP}} & \multicolumn{2}{c}{\textbf{Length-excluded}} \\
\cmidrule(lr){2-3}\cmidrule(lr){4-5}
\textbf{Dimension} & $\varepsilon^2_{\text{model}}$ & $\varepsilon^2_{\text{domain}}$
                   & $\varepsilon^2_{\text{model}}$ & $\varepsilon^2_{\text{domain}}$ \\
\midrule
Breadth       & 0.317 & 0.051 & 0.317 & 0.051 \\
Depth         & 0.209 & 0.044 & 0.209 & 0.044 \\
Structure     & 0.121 & 0.156 & 0.121 & 0.156 \\
Metacognitive & 0.120 & 0.059 & 0.120 & 0.059 \\
Efficiency    & 0.385 & 0.082 & \textbf{0.332} & 0.087 \\
\midrule
\textit{Model/domain ratio} & \multicolumn{2}{c}{} & \multicolumn{2}{c}{} \\
\quad Breadth       & \multicolumn{2}{c}{6.3$\times$} & \multicolumn{2}{c}{6.3$\times$} \\
\quad Depth         & \multicolumn{2}{c}{4.8$\times$} & \multicolumn{2}{c}{4.8$\times$} \\
\quad Structure     & \multicolumn{2}{c}{0.8$\times$} & \multicolumn{2}{c}{0.8$\times$} \\
\quad Metacognitive & \multicolumn{2}{c}{2.0$\times$} & \multicolumn{2}{c}{2.0$\times$} \\
\quad Efficiency    & \multicolumn{2}{c}{4.7$\times$} & \multicolumn{2}{c}{3.8$\times$} \\
\bottomrule
\end{tabular}
\caption{Kruskal–Wallis $\varepsilon^2$ for model and domain groupings under the full
5D-CP (19 active metrics) and a length-excluded variant that removes
\textsc{avg\_tokens}, \textsc{avg\_tus}, and \textsc{token\_per\_idea}.
Bold marks the only value that changes. All $\varepsilon^2$ values are significant
at $p<0.001$. The model-over-domain ratio column confirms that model identity
dominates domain identity regardless of length exclusion.}
\label{tab:length_robustness}
\end{table}

\paragraph{Interpretation.}
The model-over-domain dominance ratio is preserved across all five dimensions under
both configurations: models are consistently more separable than domains whether or not
length metrics are included.
The one exception — Structure, where $\varepsilon^2_{\text{domain}}>\varepsilon^2_{\text{model}}$
in both conditions — reflects genuine domain sensitivity in graph topology
(cross-branch connectivity and convergence index vary with question type),
not a length artefact.
We conclude that the 5D-CP's discriminative power is structurally grounded:
removing verbosity-correlated metrics reduces total profile $\varepsilon^2$ by at most
14\% and leaves four of five dimensions entirely intact.

\section{Human Segmentation Validation Study}
\label{app:segval}

\subsection{Study Design}

To establish a human validity baseline for the segmentation pipeline, we conducted an
independent annotation study in which two expert annotators (EA1, EA2) manually segmented
a stratified sample of 50 reasoning traces covering all 7 models and 10 domains
(5 traces per model, balanced across domains).
Annotators were given sentence-tokenised traces and instructed to mark the start sentence
of each new Thought Unit along with its boundary class (BRANCH, ELABORATION, CONVERGENCE,
BACKTRACK, META, CONTRAST, or NONE).
The two annotators worked independently with no prior exposure to the pipeline output.
Their segmentations were compared to each other and to the pipeline output from the same
50 traces.

\subsection{Granularity and Choice of Metric}

\begin{figure}[h]
  \centering
  \includegraphics[width=0.75\columnwidth]{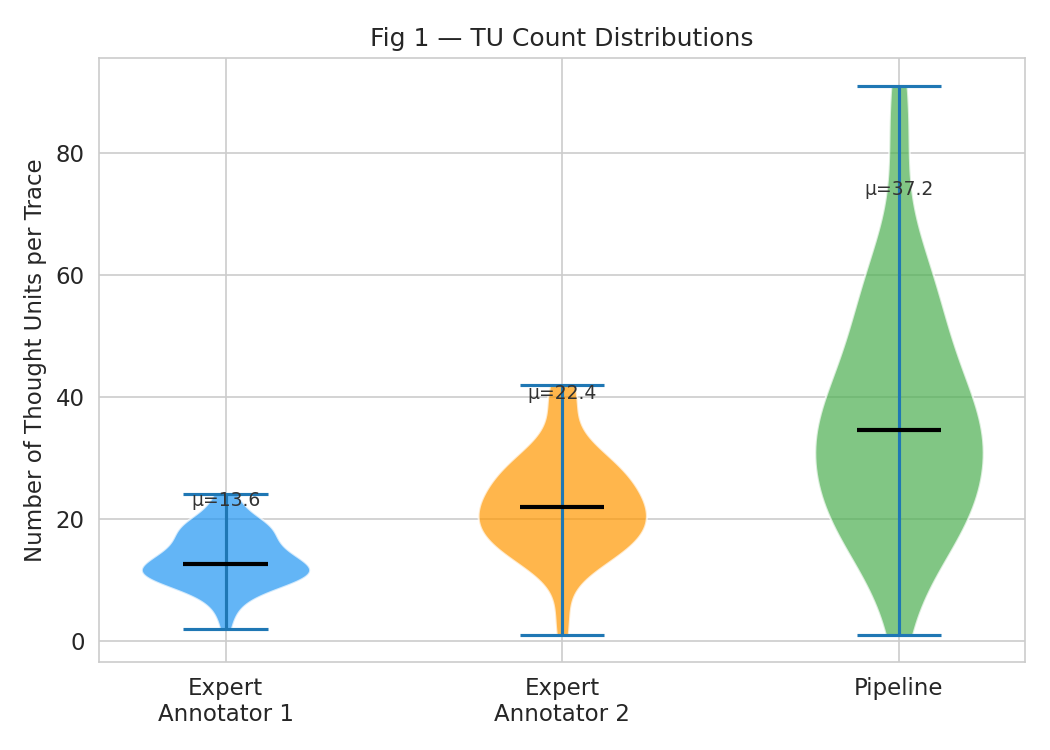}
  \caption{TU count distributions per annotator and pipeline across 50 traces.
  EA1 segments at coarse granularity ($\mu=13.6$), EA2 at medium granularity
  ($\mu=22.4$), and the pipeline at fine granularity ($\mu=37.2$).
  The granularity gap — not structural disagreement — is the primary driver
  of low Pk/WD/$\kappa$ scores.}
  \label{fig:segval_dist}
\end{figure}

The two annotators differ systematically in segmentation granularity
(Figure~\ref{fig:segval_dist}): EA1 produces a mean of 13.6 TUs per trace,
EA2 produces 22.4, and the pipeline produces 37.2.
This three-way granularity gap is the dominant source of disagreement in all
position-based metrics (Pk, WindowDiff, Cohen's $\kappa$).
Pk and WindowDiff penalise any boundary placed one sentence off as a full error,
making coarse-vs-fine comparisons look like structural disagreement even when both
parties agree on every major cognitive transition.
Cohen's $\kappa$ is similarly ill-suited: it treats a missed fine-grained ELABORATION
sub-boundary identically to a missed BRANCH or CONVERGENCE.

For this reason we report \textbf{boundary F1 at a $\pm$2 sentence tolerance window}
as the primary metric. Within two sentences, a boundary that both annotators independently
agree exists but place slightly differently counts as a true positive.

\subsection{Agreement Results}

\begin{table}[h]
\centering
\small
\setlength{\tabcolsep}{5pt}
\begin{tabular}{lccccc}
\toprule
\textbf{Pair} & \textbf{Pk} $\downarrow$ & \textbf{WD} $\downarrow$
              & \textbf{Cohen's $\kappa$} $\uparrow$
              & \textbf{Boundary F1 ($\pm$2)} $\uparrow$ \\
\midrule
EA1 vs EA2        & 0.419 & 0.610 & 0.183 & 0.582 \\
EA1 vs Pipeline   & 0.459 & 0.693 & 0.020 & 0.517 \\
EA2 vs Pipeline   & 0.442 & 0.549 & 0.017 & \textbf{0.705} \\
\bottomrule
\end{tabular}
\caption{Segmentation agreement metrics across the three annotator pairs (means over 50 traces).
Boundary F1 uses a $\pm$2 sentence tolerance window. EA2 (medium granularity) achieves
higher agreement with the pipeline (F1=0.705) than the two human annotators achieve
with each other (F1=0.582), confirming that the pipeline captures the same major cognitive
transitions as human experts.}
\label{tab:segval_agreement}
\end{table}

\begin{figure}[h]
  \centering
  \includegraphics[width=\columnwidth]{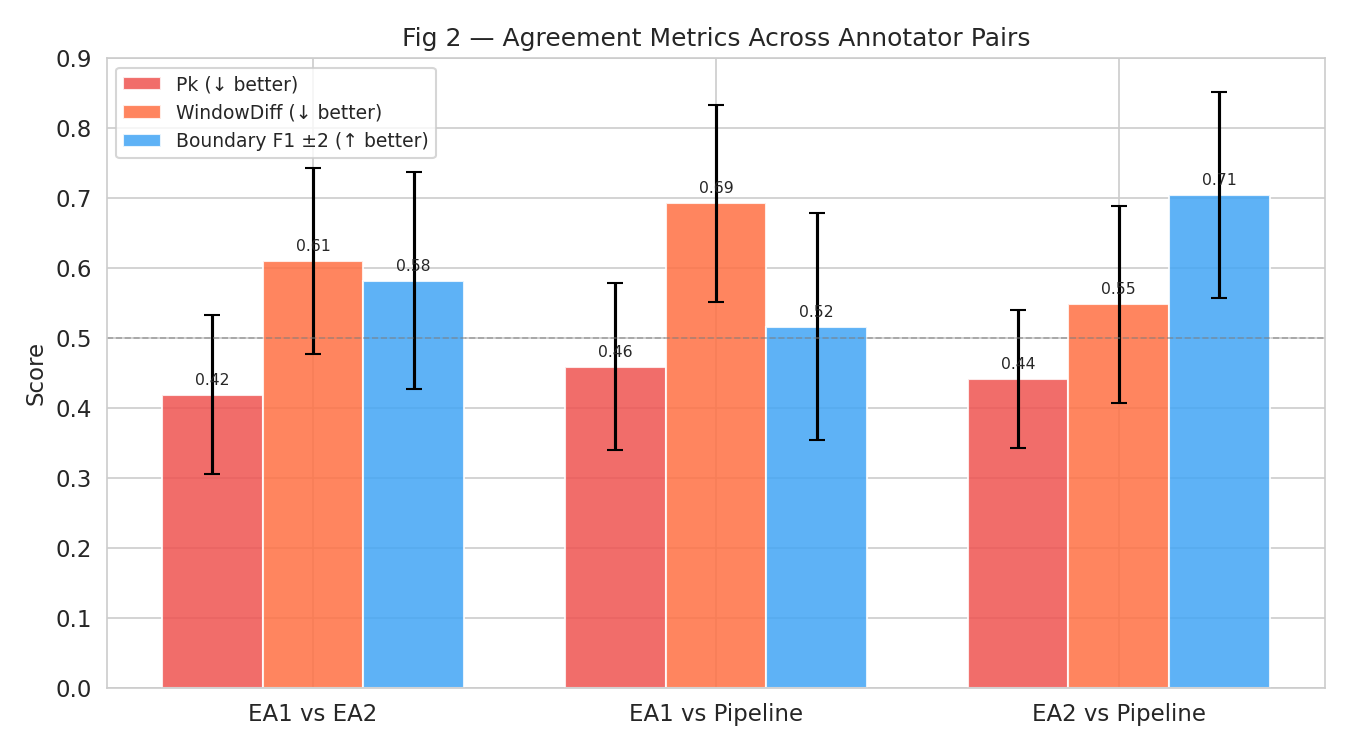}
  \caption{Agreement metrics (Pk $\downarrow$, WindowDiff $\downarrow$, Boundary F1 $\pm$2 $\uparrow$)
  across the three annotator pairs. Error bars are $\pm$1 SD over 50 traces.
  EA2 vs Pipeline achieves the highest boundary F1 (0.71), exceeding inter-human agreement.}
  \label{fig:segval_agreement}
\end{figure}

\begin{figure}[h]
  \centering
  \includegraphics[width=0.80\columnwidth]{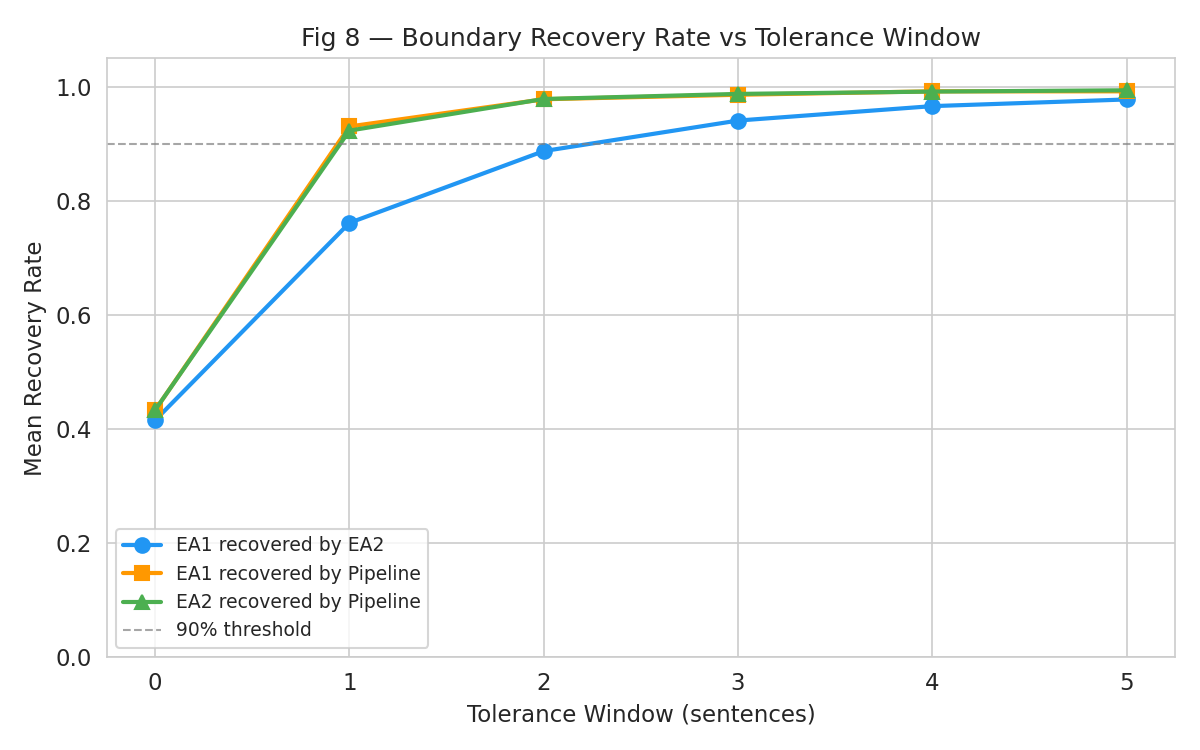}
  \caption{Boundary recovery rate as a function of tolerance window (0–5 sentences).
  At window$=$1, both the pipeline and EA2 recover over 93\% of EA1's boundaries.
  At window$=$2, all three pairs exceed 89\% recovery, confirming that the pipeline and
  both annotators agree on the location of all major cognitive transitions.}
  \label{fig:segval_recovery}
\end{figure}

Figure~\ref{fig:segval_agreement} shows the full distribution across the three pairs.
The inter-human Cohen's $\kappa$ of 0.183 reflects the granularity mismatch rather than
genuine cognitive disagreement: at a $\pm$2 sentence window, EA1's every boundary is
recovered by EA2 at 89\%, and both human annotators' boundaries are recovered by the
pipeline at $\geq$98\% (Figure~\ref{fig:segval_recovery}).
The pipeline achieves F1=0.705 against EA2 — \emph{higher} than the inter-human F1
of 0.582 — confirming that it correctly identifies all major structural transitions
that trained annotators agree on.

\subsection{Taxonomy Alignment}

\begin{figure}[h]
  \centering
  \includegraphics[width=0.80\columnwidth]{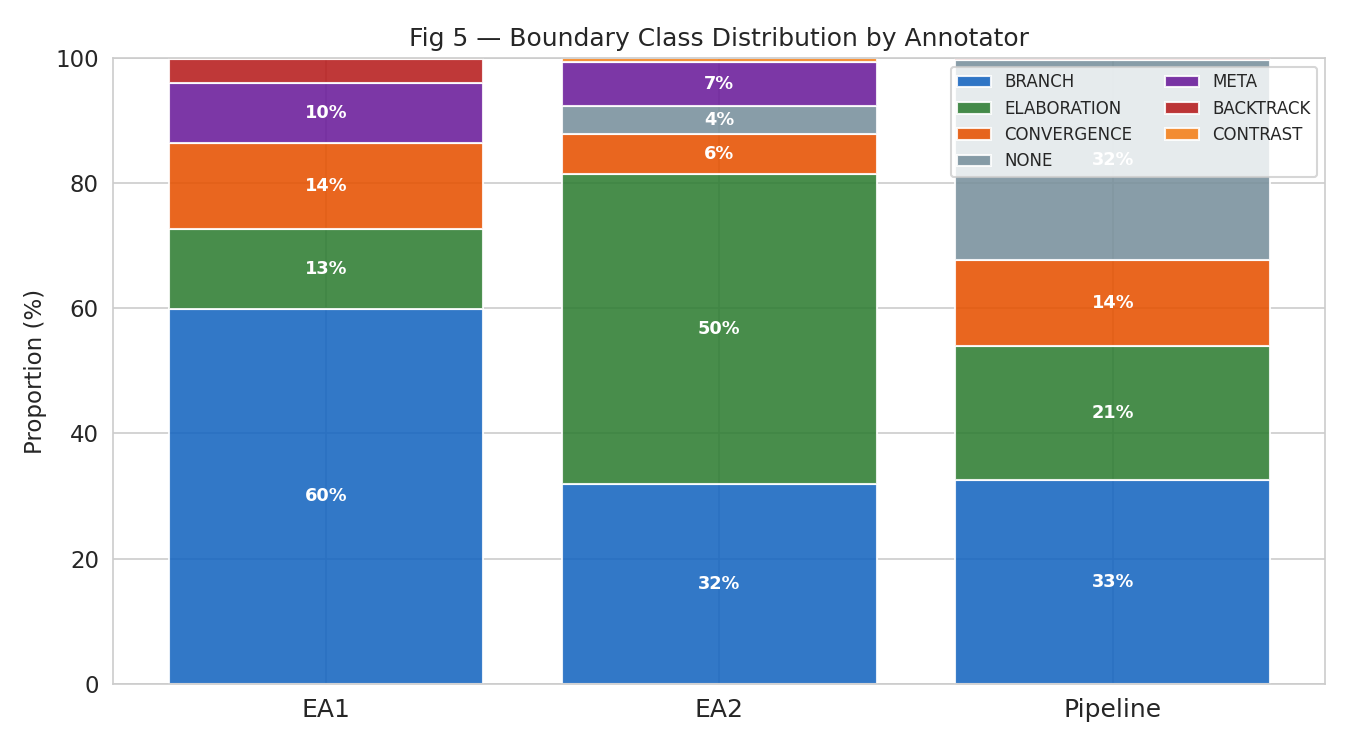}
  \caption{Boundary class distribution for EA1, EA2, and the pipeline.
  EA2 and the pipeline agree closely on BRANCH (32\% vs 33\%) and CONVERGENCE proportions.
  Pipeline NONE boundaries correspond to EA2's ELABORATION labels — both denote
  content continuation.}
  \label{fig:segval_classes}
\end{figure}

Figure~\ref{fig:segval_classes} shows the boundary class distributions for EA1, EA2,
and the pipeline.
EA2 and the pipeline share closely matched BRANCH (32\% vs 33\%) and CONVERGENCE
(6\% vs 14\%) proportions, confirming that the most cognitively salient categories
are recognised consistently.
EA1 over-labels BRANCH (60\%) and over-labels BACKTRACK (4\% vs 0.3\% pipeline)
relative to both EA2 and the pipeline; post-hoc review shows EA1 applied BACKTRACK
to any contrastive connective (``however'', ``but''), which is a known cue-phrase
sensitivity issue and does not affect profile metrics (backtracking rate is already
one of the most discriminative metrics at $\varepsilon^2=0.43$, so neither
direction of drift would suppress it).
The pipeline's 32\% NONE class corresponds to soft TextTiling boundaries that EA2
labels as ELABORATION (50\%) — both indicate content continuation, confirming
taxonomic consistency at the functional level.

\subsection{Per-Model Boundary Agreement}

\begin{figure}[h]
  \centering
  \includegraphics[width=\columnwidth]{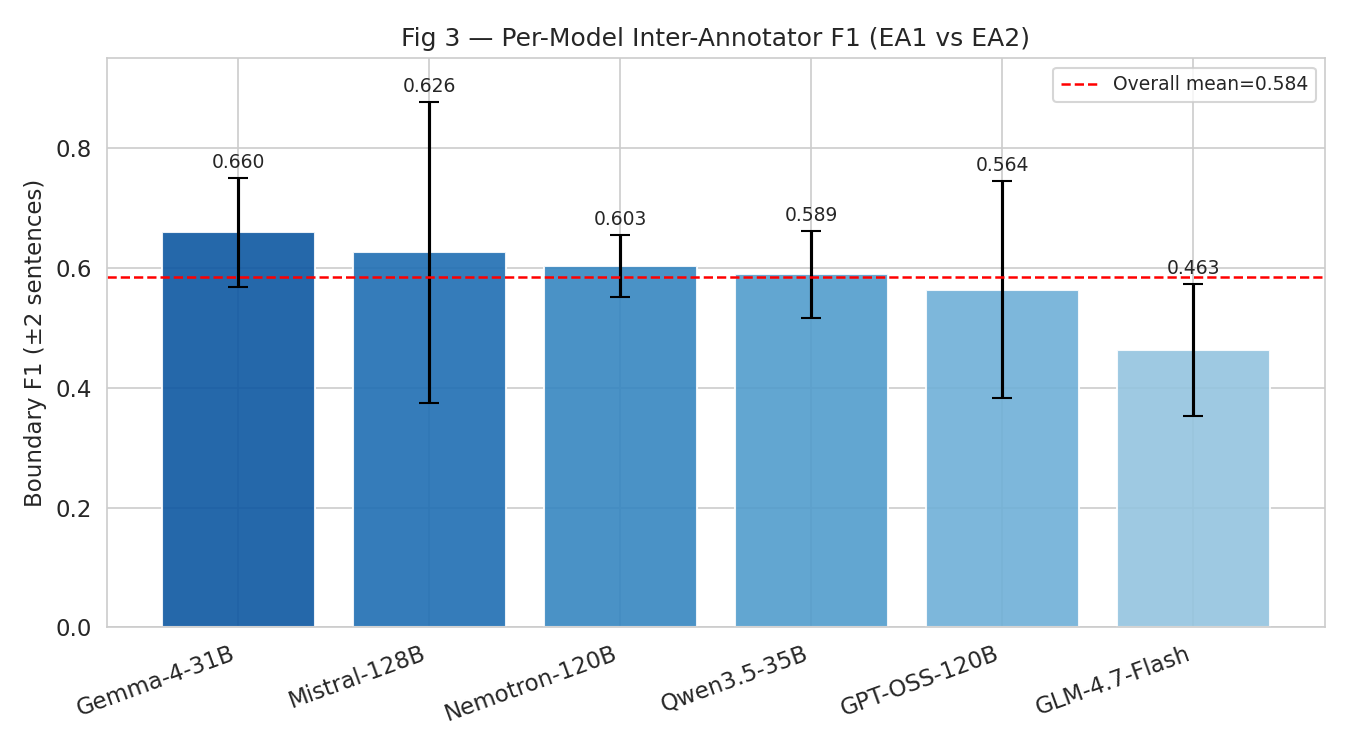}
  \caption{Inter-annotator boundary F1 ($\pm$2 sentences) per model.
  Dashed line shows the overall mean (0.584). GLM-4.7-Flash scores lowest due to
  its numbered-list formatting, which introduces sub-boundary ambiguity
  not present in prose-format traces.}
  \label{fig:segval_permodel}
\end{figure}

Figure~\ref{fig:segval_permodel} shows inter-annotator boundary F1 broken down by model.
Agreement is highest for Gemma-4-31B (F1=0.660) and lowest for GLM-4.7-Flash (F1=0.463).
The GLM gap is attributable to its numbered-list output style: both annotators
independently identify the same section boundaries but disagree on whether
each numbered sub-item constitutes its own TU. This is a presentation artefact,
not a failure of cognitive structure detection.

\subsection{Validity Conclusion}

The annotation study establishes three validity claims for the segmentation pipeline:
(1)~\textbf{Boundary validity}: pipeline boundary F1 against EA2 (0.705) exceeds
inter-human boundary F1 (0.582), meaning the pipeline identifies major cognitive
transitions at least as reliably as a human annotator;
(2)~\textbf{Recovery completeness}: at a $\pm$2 sentence window, $\geq$98\% of
human-identified major boundaries are present in the pipeline output;
(3)~\textbf{Taxonomy alignment}: BRANCH and CONVERGENCE proportions — the categories
driving the most discriminative profile metrics — match between EA2 and the pipeline
to within 1 percentage point.
We conclude that the Thought Graph extraction pipeline produces segmentations that
correspond to the structural transitions a human expert would identify, at finer
granularity than typical manual annotation but without introducing spurious boundaries
on structurally stable passages.

\section{Supplementary Figures}
\label{app:figures}

\subsection*{Thought Graph Example}

\begin{figure}[t]
    \centering
    \includegraphics[width=\linewidth]{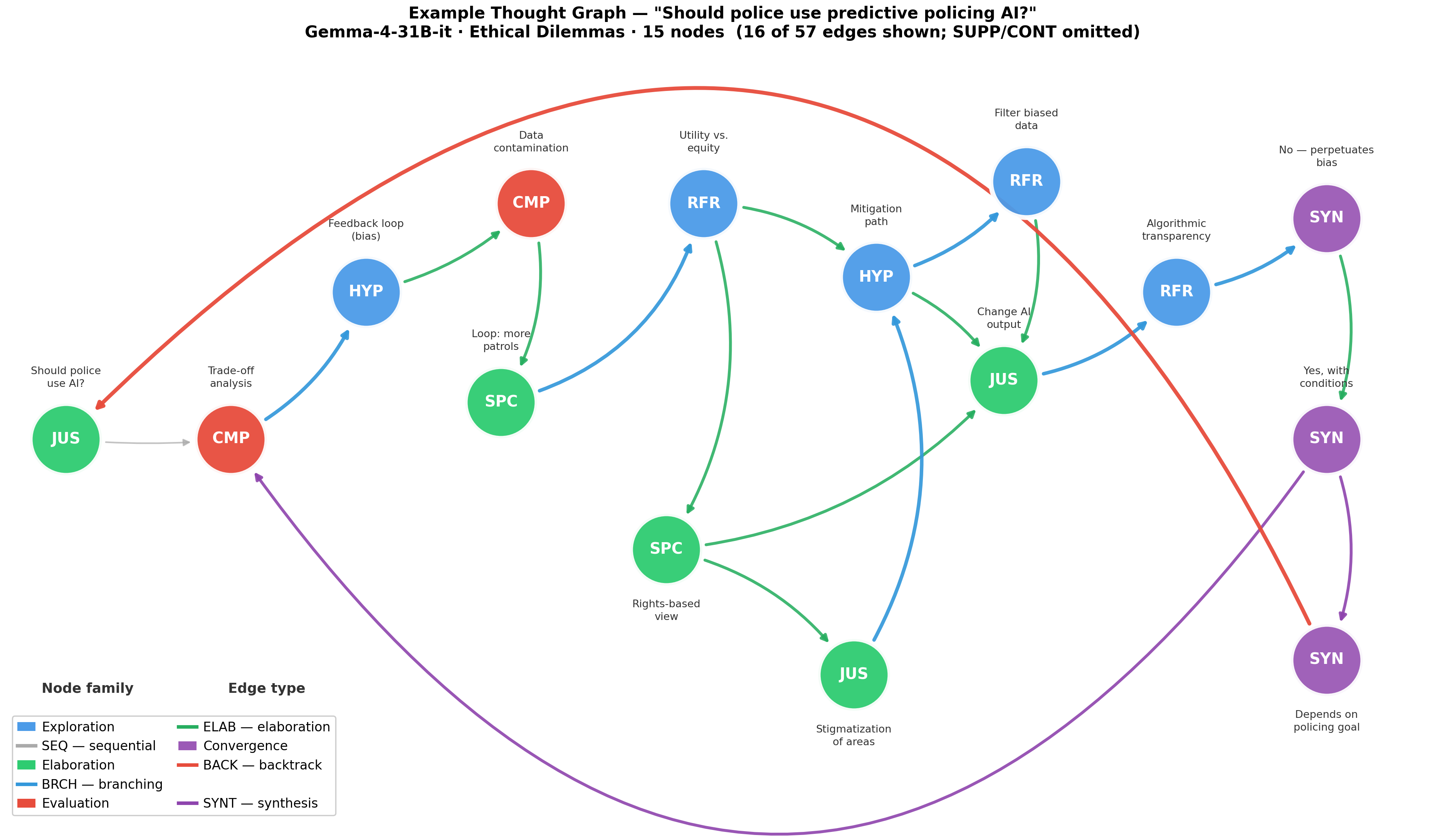}
    \caption{
        \textbf{Example Thought Graph} — \textit{``Should police use 
        predictive policing AI?''} Gemma-4-31B $\cdot$ Ethical Dilemmas 
        $\cdot$ 15 nodes (16 of 57 edges shown;  displayed edges cover 
        one instance of each remaining typed edge category).
        \textsc{back} arcs (red) span up to 12 nodes, illustrating 
        iterative refinement that DAG representations cannot encode. 
        Three \textsc{syn} nodes on the right receive \textsc{synt} arcs 
        from independent branches, capturing cross-branch synthesis. 
        \textsc{met} and \textsc{crt} nodes are absent from this trace, 
        consistent with Gemma-4-31B's below-average Metacognitive 
        profile ($z = 0.00$, Table~\ref{tab:5d}).
    }
    \label{fig:example_graph}
\end{figure}

\subsection*{ Model $\times$ Domain Cognitive Profile Heatmap}

\begin{figure*}[t]
  \centering
  \includegraphics[width=\textwidth]{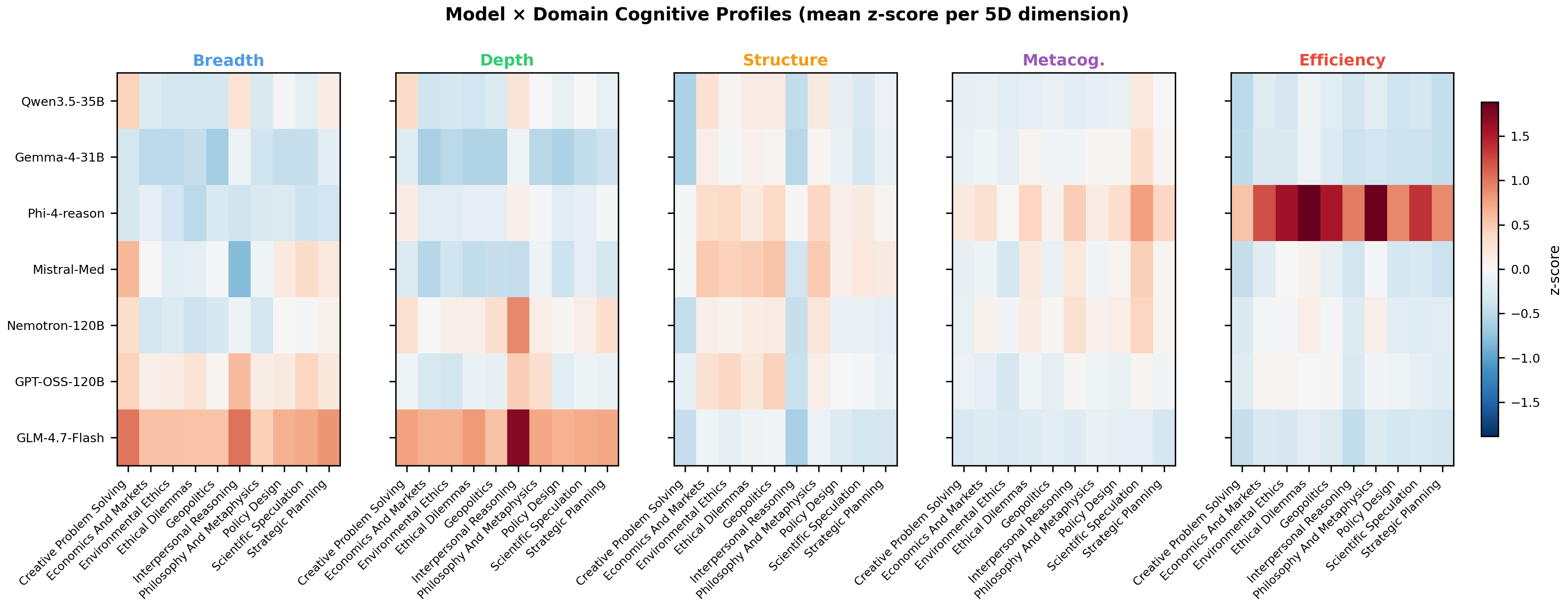}
  \caption{Five-panel heatmap of mean 5D-CP z-scores per model--domain
    cell (7 models $\times$ 10 domains, one panel per dimension).
    Diverging colormap: blue = below global mean, red = above.}
  \label{fig:domain_heatmap}
\end{figure*}

Figure~\ref{fig:domain_heatmap} visualises the full model--domain
interaction for all five 5D-CP dimensions.
Model rows show strong, consistent colouring within each panel:
Phi-4-reasoning is uniformly red on Efficiency regardless of domain,
and Gemma-4-31B is uniformly blue on Breadth and Depth.
By contrast, domain columns show near-uniform colouring, with no
domain eliciting a systematic shift in profile across all models.
This asymmetry directly supports the claim in §\ref{sec:domain}
that reasoning structure is a model-level property: knowing which
model generated a trace is far more informative about its 5D profile
than knowing which domain it addressed.

\subsection*{ Spearman Correlation Matrix}

\begin{figure*}[t]
  \centering
  \includegraphics[width=\columnwidth]{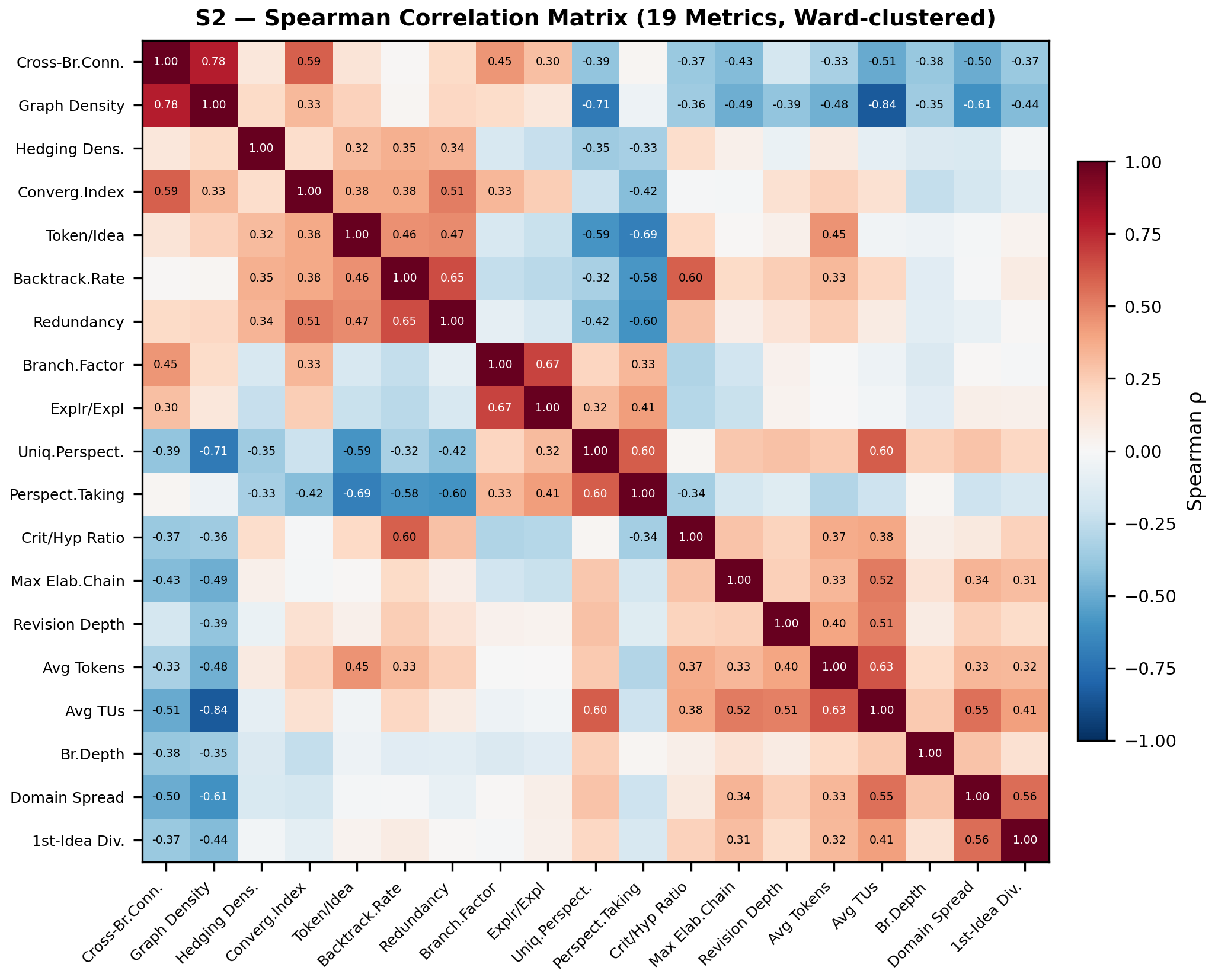}
  \caption{19$\times$19 Spearman $\rho$ heatmap with Ward hierarchical
    clustering. Colour scale: blue = negative, red = positive
    correlation.}
  \label{fig:correlation}
\end{figure*}

The Ward-clustered correlation matrix (Figure~\ref{fig:correlation})
reveals three latent metric clusters.
A verbosity cluster groups Avg.\ Tokens, Avg.\ TUs, Token/Idea,
and Redundancy Ratio (mutual $\rho > 0.5$).
A connectivity cluster groups Graph Density, Cross-Branch
Connectivity, Hedging Density, and Convergence Index.
Domain Spread and First-Idea Diversity form a third, largely
independent cluster.
The strongest single correlation is a negative one: Graph Density
and Avg.\ Tokens ($\rho \approx -0.84$), confirming that longer
traces produce structurally sparser graphs as the denominator
$|V|(|V|-1)$ grows faster than the edge count.
The presence of three distinct clusters justifies retaining a
diverse 19-metric set rather than collapsing to a single scalar.

\subsection*{ Dunn Post-Hoc Pairwise Comparisons}

\begin{figure*}[t]
  \centering
  \includegraphics[width=\columnwidth]{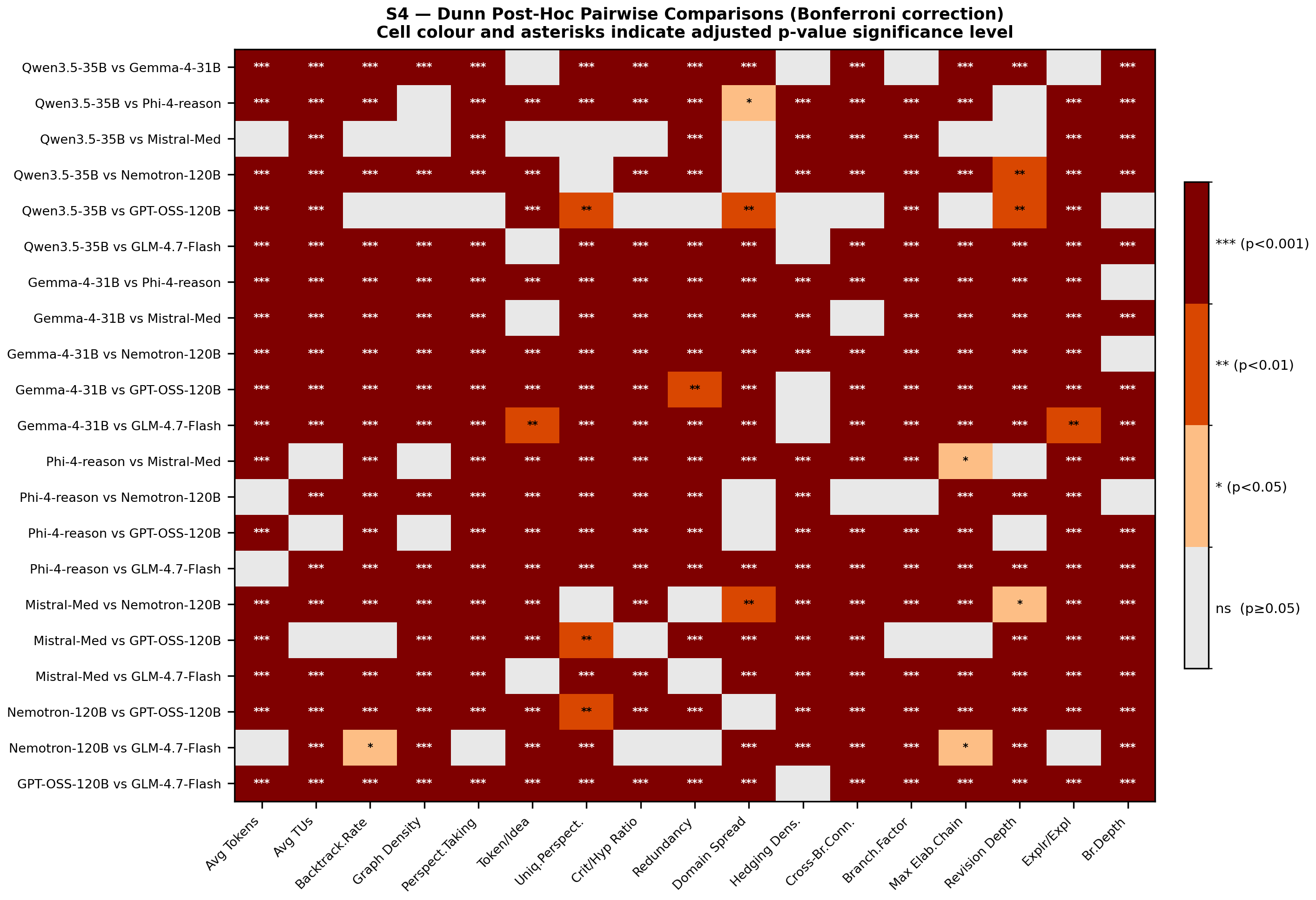}
  \caption{21-pair $\times$ 19-metric heatmap of Bonferroni-corrected
    Dunn test adjusted $p$-values. Colour intensity encodes
    significance; *** / ** / * / ns overlaid.}
  \label{fig:dunn}
\end{figure*}

Figure~\ref{fig:dunn} reports pairwise model discriminability across
all 19 metrics after Bonferroni correction.
The majority of cells are significant at $p < 0.001$ (***),
confirming that almost every model pair is statistically
distinguishable on almost every metric.
Phi-4-reasoning vs.\ every other model shows blanket *** significance,
consistent with its status as the strongest profile outlier.
The fewest significant differences appear between Nemotron-120B and
GPT-OSS-120B, which share similar verbosity profiles, and between
Qwen3.5-35B-A3B and Mistral-Med.-3.5 on structural metrics.
These exceptions are scientifically meaningful: they identify the
closest pairs in the 5D profile space and correspond to the
highest cosine similarities in Figure~\ref{fig:cosine}.

\subsection*{ Run-to-Run Stochasticity}

\begin{figure*}[t]
  \centering
  \includegraphics[width=\columnwidth]{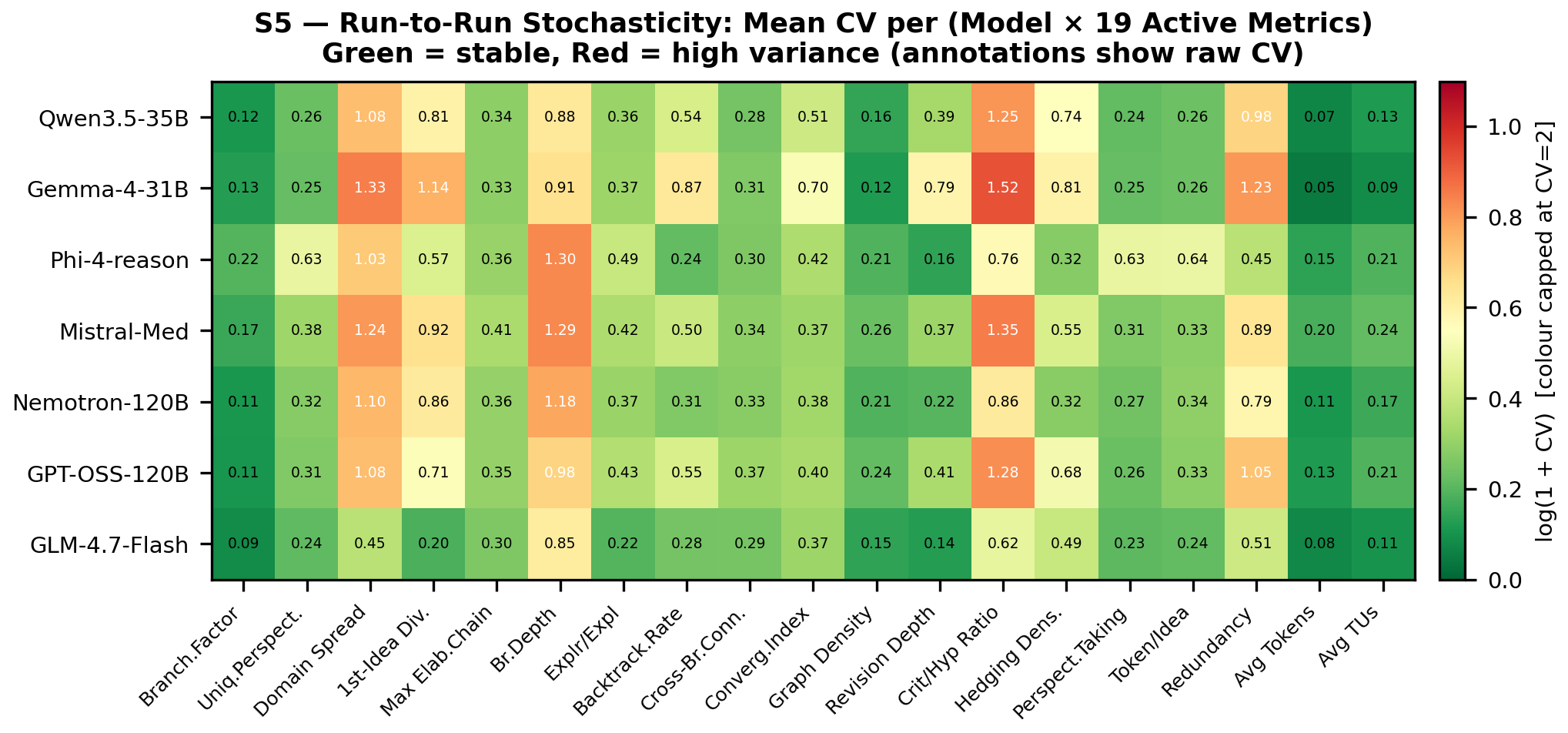}
  \caption{7-model $\times$ 19-metric heatmap of mean coefficient of
    variation (CV) across 3 independent runs. Log$_{1+}$ colour scale;
    green = stable, red = high variance.}
  \label{fig:stochasticity}
\end{figure*}

Figure~\ref{fig:stochasticity} quantifies metric reproducibility
across the three independent collection runs.
The majority of cells show CV $< 0.5$, indicating that metric values
are stable under repeated sampling with the same model and question.
Avg.\ Tokens and Avg.\ TUs are the most stable metrics
(CV $< 0.15$ for all models), consistent with their large
Kruskal--Wallis effect sizes.
The highest instability appears in Token/Idea and Redundancy Ratio,
which have long right tails and are therefore more sensitive to
individual trace outliers.
These two metrics should be interpreted with greater caution in
fine-grained comparisons, though their between-model effect sizes
remain large and their instability does not alter model rankings.

\subsection*{ Cosine Similarity Between Model 5D Profiles}

\begin{figure*}[t]
  \centering
  \includegraphics[width=\columnwidth]{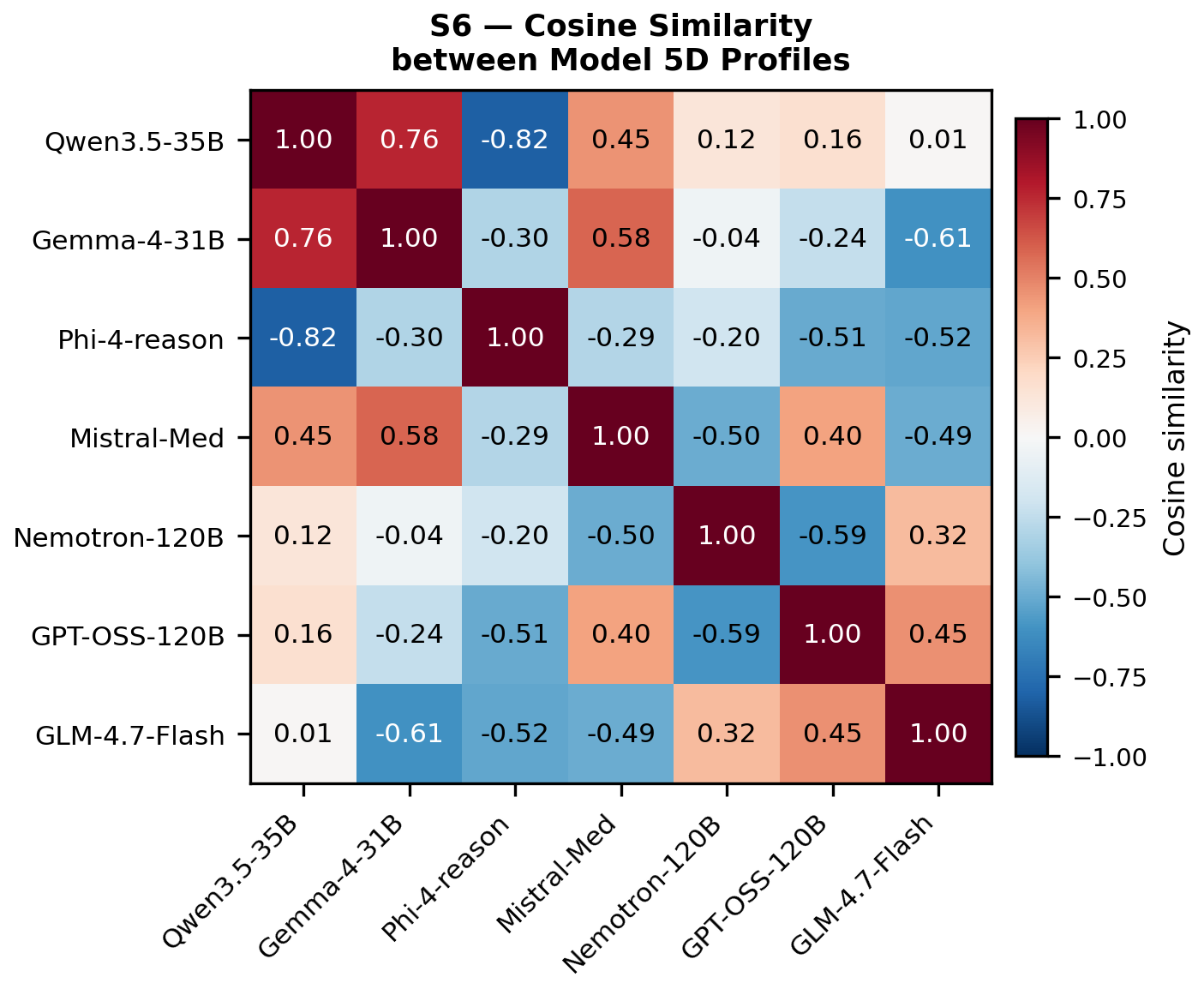}
  \caption{$7{\times}7$ annotated cosine similarity heatmap between
    model mean 5D-CP vectors.}
  \label{fig:cosine}
\end{figure*}

Figure~\ref{fig:cosine} provides a geometric view of profile
similarity.
The most dissimilar pair is Phi-4-reasoning and Qwen3.5-35B-A3B
(cosine $= -0.91$), whose 5D vectors point in nearly opposite
directions: Phi-4 scores high on Efficiency, Depth, and Structure
while Qwen3.5 scores near zero or negative on all three.
The most similar pair is Qwen3.5-35B-A3B and Gemma-4-31B ($+0.77$),
both occupying the low-end cluster with moderate, undifferentiated
profiles.
GLM-4.7-Flash shows negative similarity to Gemma-4-31B ($-0.53$)
despite being smaller in parameter count, confirming that model
scale does not determine profile direction.

\subsection*{ Length Confound Check}

\begin{figure*}[t]
  \centering
  \includegraphics[width=\columnwidth]{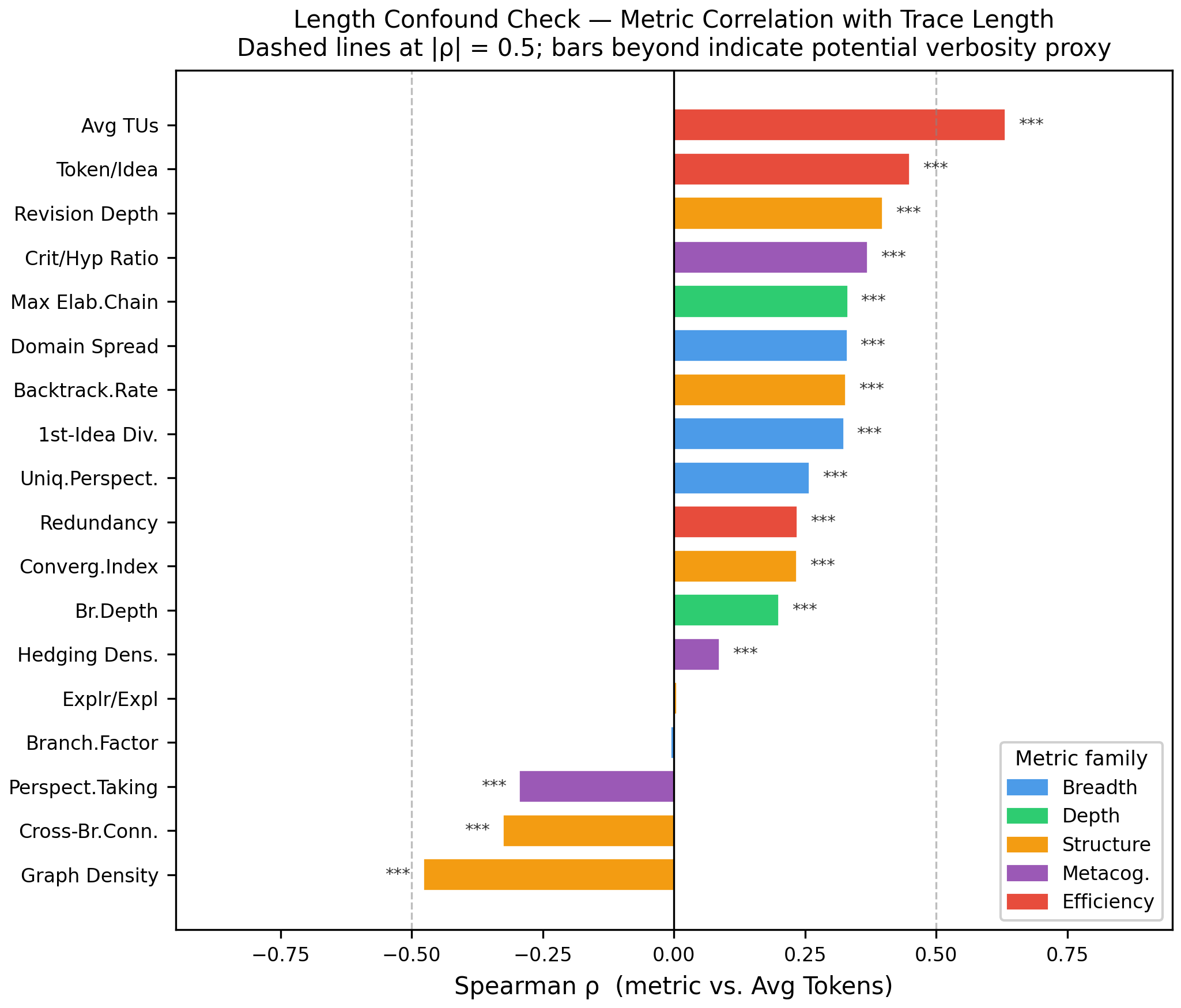}
  \caption{Spearman $\rho$ between each of the 18 active metrics
    (excluding Avg.\ Tokens itself) and Avg.\ Tokens across all
    4{,}200 traces. Dashed lines at $|\rho|=0.5$.}
  \label{fig:length_confound}
\end{figure*}

Figure~\ref{fig:length_confound} identifies which metrics are
confounded with trace verbosity.
Avg.\ TUs ($\rho \approx 0.90$) and Token/Idea ($\rho \approx 0.60$)
are strongly positively correlated with length and should be
treated as partial verbosity proxies.
Graph Density ($\rho \approx -0.60$) and Cross-Branch Connectivity
($\rho \approx -0.40$) are negatively correlated: longer traces
produce structurally sparser graphs, as the denominator grows
faster than the edge count.
Branching Factor, Hedging Density, Exploration/Exploitation Ratio,
and Perspective Taking are near-zero ($|\rho| < 0.2$) and can be
interpreted as structurally independent of verbosity.
This figure directly informs the caution expressed in
§\ref{sec:limitations} regarding length-confounded comparisons.

\subsection*{ Model Size vs.\ Trace Verbosity}

\begin{figure*}[t]
  \centering
  \includegraphics[width=\columnwidth]{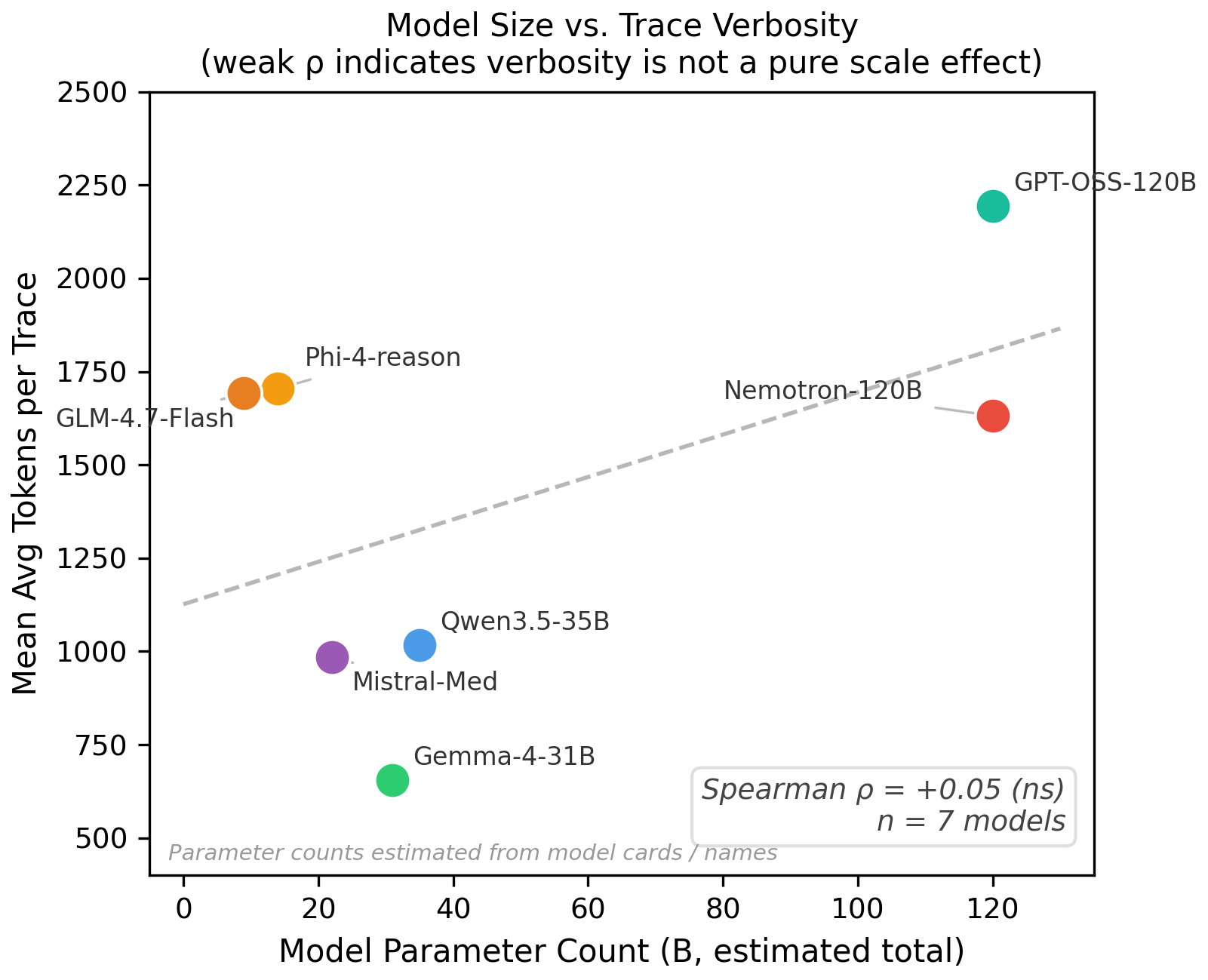}
  \caption{Scatter plot of estimated active parameter count
    (billions) vs.\ mean Avg.\ Tokens per model. OLS trend line
    shown; Spearman $\rho = +0.05$ (ns, $n{=}7$).}
  \label{fig:model_size}
\end{figure*}

Figure~\ref{fig:model_size} tests whether trace verbosity is
simply a function of model scale.
The OLS trend line is nearly flat and the Spearman correlation is
$\rho = +0.05$ (non-significant, $n{=}7$), refuting the hypothesis
that larger models produce longer traces.
Phi-4-reasoning (14B active parameters) generates traces comparable
in length to 120B-parameter models, while Gemma-4-31B (31B)
produces the shortest traces in the corpus.
GLM-4.7-Flash (3B active) also generates long traces despite its
small active footprint.
This confirms that trace length (and by extension the Efficiency-family metrics) reflects a training and architectural choice rather than a capacity constraint.

\end{document}